\documentclass[lettersize,journal]{IEEEtran}

\usepackage{amssymb,amsfonts,amsthm}
\usepackage{amsmath}
\usepackage{algorithmic}
\usepackage[ruled,linesnumbered]{algorithm2e}
\usepackage{graphicx}
\usepackage{cite}
\usepackage{textcomp}
\usepackage[labelformat=simple]{subcaption}

\usepackage{fancyhdr}

\usepackage{psfrag}
\usepackage {epstopdf}
 
\usepackage[hidelinks]{hyperref}
\usepackage{multirow}

\usepackage{float}
\usepackage[utf8]{inputenc}
\usepackage{tabularx}  
\usepackage{titlesec}

\usepackage{xcolor}
\usepackage{authblk}
\usepackage{caption}

\usepackage[export]{adjustbox}
\usepackage[hang,flushmargin]{footmisc}
\usepackage{bm}
\def\BibTeX{{\rm B\kern-.05em{\sc i\kern-.025em b}\kern-.08em
    T\kern-.1667em\lower.7ex\hbox{E}\kern-.125emX}}

\title{Resource Allocation of Federated Learning for the Metaverse with Mobile Augmented Reality} 
\date{}

\usepackage{graphicx}
\usepackage{mathrsfs}

\newtheorem{theorem}{Theorem}
\newtheorem{lemma}{Lemma}

\newcommand{\junzhao}[1]{\iffalse\ding{110}\ding{43}\fi\textcolor{black}{Jun Zhao: #1}}

\begin{document}

\author{Xinyu~Zhou,~Chang~Liu, Jun Zhao
\thanks{A shorter version of this work containing fewer results has been accepted to 2022 IEEE 42nd International Conference on Distributed Computing Systems (ICDCS)~\cite{ICDCS}. Improvements over~\cite{ICDCS} are explicitly discussed in this paper.}
\thanks{The authors are all with the School of Computer Science and Engineering, Nanyang Technological University, Singapore. Xinyu Zhou and Chang Liu are also with ERI@N, Interdisciplinary Graduate Programme, Nanyang Technological University, Singapore. Contact: xinyu003@e.ntu.edu.sg, liuc0063@e.ntu.edu.sg, JunZHAO@ntu.edu.sg
}  
\thanks{This research is partly supported by the Singapore Ministry of Education Academic Research Fund under Grant Tier 1 RG90/22, Grant Tier 1 RG97/20, Grant Tier 1 RG24/20 and Grant Tier 2 MOE2019-T2-1-176; and partly by the Nanyang Technological University (NTU)-Wallenberg AI, Autonomous Systems and Software Program (WASP) Joint Project.}
}

\maketitle
 \thispagestyle{fancy}
\pagestyle{fancy}
\lhead{This paper appears in IEEE Transactions on Wireless Communications. DOI: \href{https://doi.org/10.1109/TWC.2023.3326884}{https://doi.org/10.1109/TWC.2023.3326884} }
\rhead{}
\cfoot{\thepage}
\renewcommand{\headrulewidth}{0.4pt}
\renewcommand{\footrulewidth}{0pt}

\begin{abstract}
The Metaverse has received much attention recently. Metaverse applications via mobile augmented reality (MAR) require rapid and accurate object detection to mix digital data with the real world. Federated learning (FL) is an intriguing distributed machine learning approach due to its privacy-preserving characteristics. Due to privacy concerns and the limited computation resources on mobile devices, we incorporate FL into MAR systems of the Metaverse to train a model cooperatively. 
Besides, to balance the trade-off between energy, execution latency and model accuracy, thereby accommodating different demands and application scenarios, we formulate an optimization problem to minimize a weighted combination of total energy consumption, completion time and model accuracy.
Through decomposing the \mbox{non-convex} optimization problem into two subproblems, we devise a resource allocation algorithm to determine the bandwidth allocation, transmission power, CPU frequency and video frame resolution for each participating device. We further present the convergence analysis and computational complexity of the proposed algorithm.
Numerical results show that our proposed algorithm has better performance (in terms of energy consumption, completion time and model accuracy) under different weight parameters compared to existing benchmarks. 
\end{abstract}


\section{Introduction}

\subsection{\textcolor{black}{Background}}
Nowadays, we can develop intelligent machine learning models that reliably recognize and classify complicated things in the real world due to advances in deep learning. This breakthrough has motivated the advancement of Augmented Reality (AR) technology, which improves the human perception of the world by merging the actual world with the virtual space. 
Mobile Augmented Reality (MAR) applications operate on mobile platforms and allow users to access services through their devices' interfaces. MAR and augmented reality (AR)  further empower the Metaverse, which is considered the next evolution of the Internet~\cite{park2022metaverse}.
By implementing machine learning technologies on devices, more intelligent MAR applications are developed in different Metaverse scenarios, such as entertainment, communications and medicine. 

Imagine a Metaverse scenario where putting on a pair of glasses, and you can see descriptions of buildings, which is convenient for finding the destination during traveling. 
Thus, in MAR for the Metaverse, rapid and accurate object recognition is critical in bringing digital data together with environmental aspects. 
This scenario can be addressed by different approaches. 
For example, mobile devices can upload their images to a proximate edge server with strong computing power to run deep learning algorithms to save computational latency and energy \cite{zhang2018jaguar, park2020deep}.

In this paper, we assume each AR user of the Metaverse can generate data for object detection tasks. For example, users can mark labels for images generated by AR devices and those images are used for training afterward. 
To create an object detection model with high accuracy, we usually need to feed it with a large size of training data. Due to limitations of user privacy, data security and competition among industries, usually, the data is limited.
Besides, the concern around data privacy is rising. Related laws and regulations are appearing, such as the General Data Protection Regulation  (GDPR) of the European Union  (EU), so it may be challenging to collect sensitive data together to learn a model. 
Mobile devices offloading their training tasks to a cloud/edge server may raise privacy concerns.
{\color{black} Training a model solely on one device could be a solution, but the amount of data collected by one device may not be enough for training a good model.}
We then consider incorporating federated learning (FL) into the design of MAR systems for the Metaverse. 

FL can collaboratively train a machine learning model while preserving data on its local device distributedly. It is a special distributed learning framework proposed by Google in \cite{mcmahan2017communication}. 
In the FL algorithm of \cite{mcmahan2017communication}, each device trains a model locally for a certain number of local iterations and periodically uploads its parameters to a server to build a shared global model. By incorporating FL, on the one hand, we can utilize the computation resources of each device. On the other hand, we do not need to upload video frames to a central server for training, which protects user privacy. We call such a system as FL-MAR system.

\subsection{\textcolor{black}{Motivation}} 
The hardware performance is improving continuously according to Moore's Law \cite{dionisio20133d}, and thus the computational resources of mobile devices will most likely become abundant in the future. 
Therefore, it is worth investigating how to run machine learning experiments on mobile devices. 
One MAR device usually has limited data, which may not be enough to train a good model solely. That is why we are considering integrating FL into MAR applications.
Each MAR device can benefit from joining FL. 
FL only needs each device to upload its parameters to a server, and the server will send back the average parameter to each device.
Thus, each device can improve its model without sharing its data, which protects user privacy.
Hence, it is worthwhile to study the application of FL to MAR for building intelligence into the Metaverse.
\textcolor{black}{Moreover, there are other promising applications of FL in the Metaverse: 1)~\underline{Personalized content recommendation}: With FL, the users have the ability to partake in services offered in the Metaverse, such as receiving visual content recommendations, while ensuring the
safeguarding of their privacy \cite{lee2021all}. 2)~\underline{Object detection}: FL could help enable users to identify and interact with visual objects in the Metaverse with the technology of mobile augmented reality (MAR). This is also the scenario we consider in this paper. 3) \underline{Training avatars}: FL could be utilized to train avatars powered by artificial intelligence to provide users with personalized experiences while protecting user privacy. Additionally, AR or virtual reality (VR) in the Metaverse is accompanied by the privacy leakage of sensitive information, such as gender, age, ethnicity, etc. \cite{nair2022exploring}. Therefore, FL can help prevent the risk of privacy leaks to a certain extent. \textcolor{black}{For more details on using FL for the Metaverse, interested readers can refer to recent survey/tutorial papers \cite{FLMetaverse2023, bashir2023survey}.
}}

\subsection{\textcolor{black}{Challenges}}
Applying FL to MAR applications for the Metaverse still has high requirements of device memory, computational capability6 and communication bandwidth.
There are several challenges to the deployment of FL to MAR applications: (1) Limited bandwidth results in long \textit{latency} between clients and the server and, consequently, affects the convergence time of the global model. (2) A large amount of \textit{energy} is needed because a satisfactory model requires quite a few local computation and communication rounds. (3) MAR devices may be in a low-battery condition while training. How to utilize its computational resources (e.g., computational frequency) and adjust video frame resolutions during training to save energy and time? 
Besides, finding optimal resource (e.g., transmission power, bandwidth, CPU frequency, etc.) allocation strategies can save both communication and computation resources \cite{yang2020energy, luo2020hfel, dinh2020federated,qian2022}.
An optimal resource allocation strategy will allow us to reduce the execution latency and energy consumption and accelerate the convergence of the global model. 
Therefore, it is essential to consider adjusting bandwidth, transmission power and CPU clock frequency for each participating device to optimize energy consumption and completion time \cite{bouzinis2021wireless}.
Moreover, different scenarios have different energy and latency demands.
For example, connected autonomous vehicles in an intelligent transportation system require low-latency control decisions, while personal computers may tolerate higher latency to utilize the idle time to train their model.
Besides, the object recognition accuracy is related to the video frame resolution of MAR devices \cite{liu2018edge}. Higher resolution means higher computational latency and energy consumption.
Thus, it raises the following question: how to assign the transmit power, bandwidth, CPU frequency and video frame resolution to each user device to find the optimal trade-off between energy consumption, the total completion time and model accuracy of the FL-MAR system while accommodating different scenarios?

Motivated by the above question, we consider a basic FL-MAR model for the Metaverse with frequency-division multiple access (FDMA) and formulate a joint optimization of energy, time and accuracy problem over the wireless network. 
The objective function includes a weighted combination of total energy consumption, completion time and model accuracy. 
The participating devices in FL are under power constraints and bandwidth constraints while pursuing as little energy, time consumption and much accuracy as possible.

\textbf{Comparison with our ICDCS paper~\cite{ICDCS}}. We extend the algorithm in \cite{ICDCS} to an MAR scenario for the Metaverse. In \cite{ICDCS}, we do not consider any optimization variable or objective function related to MAR. The optimization problem in \cite{ICDCS} is to minimize the total energy and time consumption. 
In this work, we add one more optimization variable---video frame resolution in the context of MAR for the Metaverse. The objective  is to optimize energy, time, and model accuracy simultaneously.
To evaluate the impact of video frame resolution,     object detection experiments are implemented on COCO dataset~\cite{lin2014microsoft} using the modified YOLOv5m~\cite{glenn_jocher_2022_7002879} model.

 \textbf{Contributions}. In summary, the main contributions of this paper are listed as follows:
\begin{itemize}
    \item To the best of our knowledge, we are the first to incorporate FL into the energy-efficient MAR system design for the Metaverse. To utilize the communication and computational resources efficiently, we formulate an optimization problem with variables including transmission power, bandwidth, CPU frequency and video frame resolution, with the objective function incorporating energy, time, and model accuracy.
    \item One part of the optimization problem includes a non-convex sum-of-ratio form of communication energy consumption, which cannot be solved by traditional optimization techniques. To tackle this problem, we propose to decompose the original problem into two subproblems: the first subproblem is to minimize computation energy, completion time and model accuracy, and the second is to minimize the communication energy. The first one is solved through its Lagrange dual problem. The second subproblem is a sum-of-ratio minimization problem, which is transformed into a convex subtractive-form problem through a Newton-like method similar to \cite{jong2012efficient} used for solving a general fractional programming problem. 
    \item Convergence analysis and time complexity are provided. Numerical results show: 1) Under different weight parameters (i.e., different requirements), our resource allocation algorithm can find a corresponding solution to jointly minimize energy consumption and completion time and maximize the model accuracy. 2) Our joint optimization framework's performance is better than the communication-only optimization and computation-only optimization frameworks. 3) Our resource allocation scheme outperforms the benchmarks, especially regarding total energy consumption.
\end{itemize}

In Section \ref{sec:literature_review}, we discuss the related work. In Section \ref{sec:sys_model}, we introduce the system model and necessary notations. Problem formulation is provided in Section \ref{sec:prob_formu}. We elaborate on the proposed algorithm in Section \ref{sec:solution_joint_prob}. Then Section~\ref{sec:conv} presents the algorithm's convergence and computational complexity. 
 Numerical results are shown in Section \ref{sec:experiments}. Finally, Section \ref{sec:conclusion} concludes the paper.

\section{Related Work} \label{sec:literature_review}

In this section, we survey work related to our study in terms of MAR, the Metaverse and FL.

 \subsection{\textcolor{black}{Related studies on MAR and Metaverse}}
 Recently, MAR systems and their empowerment for the Metaverse are attracting more attention thanks to the improvements in wireless communications and end devices' computing capacity~\cite{chatzopoulos2017mobile,bottani2019augmented}.
 The MAR applications, especially for the Metaverse, are CPU-intensive and memory-intensive since they perform various heavy computational tasks.
 
\textbf{\textcolor{black}{Task offloading and edge/cloud servers}}. A few studies focused on developing new methods to offload tasks to the cloud or edge server to reduce energy consumption and improve the performance of MAR systems.
For example, \cite{wang2019auto} proposed an edge-assisted scheme to schedule offloading requests across multiple applications. 
\cite{9155517} designed an edge-based MAR system and proposes an optimization algorithm to change MAR clients' configuration parameters and guide edge server radio resource allocation dynamically to save energy consumption and latency while maximizing the model accuracy.
Authors in \cite{liu2018edge} devised an edge orchestrator to complete object analysis for MAR. 
\textcolor{black}{They formulated the model accuracy as a function of the video frame resolution and conducted experiments based on YOLO algorithm \cite{redmon2016you} to construct the accuracy function concerning different video frame resolutions.}  \cite{liu2018dare} \textcolor{black}{allocated} computation resources dynamically and utilizes computation resources on the edge
server.

\textbf{\textcolor{black}{Resource allocation and mobile edge computing}}. Another genre of work aimed at studying resource allocation schemes in MAR systems with mobile edge computing \cite{lin2021resource,zhang2017live,zhou2020communication,ahn2019novel, he2020optimizing}. 
For example,~\cite{lin2021resource} studied the joint problem of viewport rendering offloading and resource allocation for wireless VR streaming.
VideoStorm~\cite{zhang2017live} jointly optimized resources and quality in the video analytics system.~\cite{zhou2020communication} designed an algorithm to minimize the maximum communication and computation latency among all VR users under the limited computation resource and transmission power.
\cite{ahn2019novel} proposed a joint resolution and power control scheme considering energy efficiency, latency and accuracy trade-off in the MAR system.\textcolor{black}{\cite{he2020optimizing} studied the communication and computation resource allocation problem for the mobile edge systems where CNN models were deployed on edge servers to support AR applications.}

\textbf{\textcolor{black}{On-device learning}}. There are also studies devising new on-device deep learning frameworks in MAR \cite{huynh2017deepmon,han2016mcdnn,jiang2021flexible,park2021enabling}.
However, an individual data holder may not have sufficient local data or features for training an effective model. In our work, we incorporate the framework of federated learning, where mobile devices are allowed to learn a shared global model collaboratively. 

Related work on the Metaverse has been growing recently \cite{kang2022blockchain,zeng2022HFedMS,xu2022wireless, si2022resource}. 
For instance, \cite{kang2022blockchain} and \cite{zeng2022HFedMS} were about the application of FL to the industrial Metaverse. \cite{xu2022wireless} stood in the perspective of economics to assess and improve VR services for the Metaverse. \cite{si2022resource} studied the resource allocation problem in a Metaverse MAR service model. However, there is still a lack of FL and MAR in the Metaverse scenario.

\subsection{\textcolor{black}{Related studies on FL}}
In FL, due to limited communication resources and numerous user devices, the convergence rate of the global model may be slow, resulting in high latency and communication costs.

\textbf{\textcolor{black}{FL and wireless communications}}. Related studies \cite{yang2020energy, dinh2020federated, luo2020hfel, zeng2020energy,liu2021joint, liu2022joint, li2021delay,9528995} paid much attention to the whole FL system and devised an approach to allocating energy resources to each device in the system. The resources related to energy consumption involve limited bandwidth, transmit power, local device's CPU frequency and so forth.
Considering all devices joining each communication round via FDMA, a joint learning and communication problem was formulated in \cite{yang2020energy}. Then, the authors investigated how to allocate time, bandwidth, CPU frequency and learning accuracy to each device under a latency constraint. Similarly, instead of considering FDMA, Dinh \emph{et al.} \cite{dinh2020federated} came up with an FL algorithm via time-division multiple access (TDMA) called FEDL and studied the trade-off between the convergence time and energy consumption of FEDL. 
\textcolor{black}{Some papers also studied user (i.e., device/edge) association to reduce energy or latency for FL. \cite{luo2020hfel} \textcolor{black}{proposed} a resource allocation and edge association scheme in a cloud-edge-client federated learning framework and \textcolor{black}{showed} it to converge to a stable system point. Resource allocation and edge-device selection policy \textcolor{black}{were} discussed in \cite{zeng2020energy}.
\cite{liu2021joint} introduced a solution to reduce learning latency in wireless federated learning by implementing model pruning and device selection. \cite{liu2022joint} examined how imbalanced data distribution affected the convergence rate and the learning accuracy and presented algorithms that combined user association and wireless resource allocation.} 
\textcolor{black}{The previous work mainly aimed at reducing the latency and energy, and some work also studied the delay distribution in FL, such as \cite{li2021delay}.}

 \textbf{Related work on FL with AR}.  
We only found one article~\cite{chen2020federated} that incorporated AR into the FL study. To obtain an accurate global model and reduce the computational and power costs for AR users,~\cite{chen2020federated} proposed to use an FL-based mobile edge computing paradigm to solve object recognition and classification problem. However, \cite{chen2020federated} only proposed an FL framework without considering how to allocate communication and computation resources in this framework.

\textbf{Novelty of our work}. 
The previously mentioned approaches, which use remote clouds to ingest images or video streams, are still time-consuming and energy-intensive.
Besides, offloading tasks to cloud servers may incur privacy leakage. 
In contrast, performing tasks locally ensures a faster experience and provides more secure privacy protection.
By incorporating the framework of FL, we allow performing model training locally on devices and periodically uploading their parameters to a base station to create a shared global model. 
Without assuming that devices and servers have sufficient bandwidth, we investigate a more practical setting where the bandwidth between devices and servers is limited.
To reduce energy consumption and latency, we propose an energy-efficient MAR system for the Metaverse, where the weighted combination of energy consumption and total completion time is optimized. 
In addition to service latency and energy consumption, learning accuracy is equally vital to Metaverse MAR systems. Thus, balancing service latency, energy consumption and learning accuracy is essential. 
Although the above studies \cite{yang2020energy, luo2020hfel, dinh2020federated} consider latency and energy consumption, they did not take machine learning accuracy into account.~\cite{liu2018edge} investigated the trade-off between the latency and analytics accuracy in edge-based MAR systems and developed an algorithm to improve the system performance, but it failed to consider the energy consumption of MAR devices.
Although \cite{9155517} considered optimizing the video frame resolution, CPU frequency and the bandwidth of each user, its scenario was indoor without incorporating FL and it defined the transmission rate by using CPU frequency which was in a different context from ours. Moreover, \cite{si2022resource, he2020optimizing} optimized the resource allocation for each user in an MAR system, but they did not incorporate FL.
In our work, we maximize the accuracy for each local device while achieving low latency and energy consumption at the same time under a wireless communication scenario.
\textcolor{black}{We would like to mention that the sum of the accuracy of the local device as a part of the optimization objective function has also been considered in recent work~\cite{liu2018edge,wang2020user} on augmented reality, but~\cite{liu2018edge,wang2020user} are not about federated learning and have simpler optimization problems compared with ours. }

\begin{figure*}[tbp]
    \centering
    \captionsetup{justification=centering}
    \includegraphics[scale=0.6]{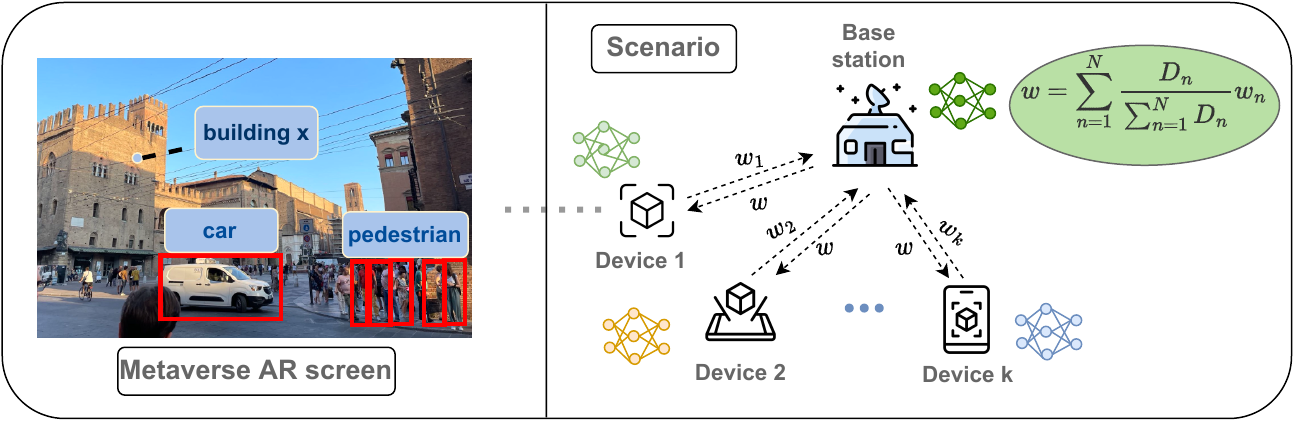}
    \caption{Federated learning with mobile augmented reality (FL-MAR) for the Metaverse.\vspace{-10pt}} 
    \label{fig:FL_framework}
\end{figure*}

\section{Metaverse System Model of FL with MAR} \label{sec:sys_model}

This section presents the Metaverse system model of FL with MAR over wireless communications. We consider a star network with a base station, as shown in Fig. \ref{fig:FL_framework}. The Metaverse service is provided through the base station.
The base station serves $N$ MAR devices from $1$ to $N$, and the set of the $N$ MAR devices is represented by $\mathcal{N} := \{1, 2, \ldots, N\}$. Let $n\in \mathcal{N}$ denote the device index. The $n$th device contains a dataset with $D_n$ samples.

As shown in Fig. \ref{fig:FL_framework}, the FL process is iterative and consists of several global rounds. 
At the beginning of each global round, each device has the same model broadcasted by the base station at the end of the previous global round. 
Each device performs training on its local dataset for several local epochs and obtains an updated model parameter. At the end of this global round, all devices upload their model parameters to the base station to compute a weighted average of the model parameter, and the base station broadcasts the new parameter to all devices to start the next global round. Such an uploading and broadcasting process is called global communication.

\textcolor{black}{Then, the overall FL model is to solve the problem $ \textstyle{\min _{\boldsymbol{w}\in\mathbb{R}^p} \ell(\boldsymbol{w}) = \sum_{n=1}^{N} \frac{D_n}{D} \ell_n(\boldsymbol{w})}$,
 where $D = \sum_{n=1}^{N} D_n$, $\textstyle{\ell_n(\bm{w})}$ is the loss function defined on each local device, and $\bm{w}$ is the global model parameter.}
Between two global aggregations, 
each device $n$ trains its own model for a few local iterations, and the aim is to minimize the loss function defined at device $n$ as $\textstyle{\ell_n(\bm{w}) = \frac{1}{D_n} \sum_{i=1}^{D_n} \ell_{n,i}(\bm{w})}.$
The function $\ell_{n,i}(\bm{w})$ measures the error between the model prediction and the ground truth label, and $i$ represents the $i$th sample on device $n$. We consider that in each local epoch,  device $n$ uses all of its $D_n$ data samples.

In this paper, we formulate a joint optimization of energy, completion time and model accuracy problem in FL via Frequency Division Multiple Access (FDMA). \textcolor{black}{FDMA means the spectral resources are divided among MAR devices. Each device is assigned a unique frequency band, which is dedicated to that user for the communication duration.} The objective function is a weighted combination of total energy, time, and model accuracy. Total energy consumption, total completion time and model accuracy model are discussed in Sections~\ref{subsec:total_e_def},~\ref{subsec:total_t_def} and~\ref{sec:Accuracy}, respectively.
Notations in this paper are listed in Table~\ref{tab:notation} in Appendix \ref{notations}. Bold symbols stand for vectors.

\subsection{Total Energy Consumption} \label{subsec:total_e_def}

The \textbf{total energy consumption}, $\mathcal{E}$, comprises \textbf{wireless transmission energy consumption} and \textbf{local computation energy consumption}. The same as many studies \cite{dinh2020federated, luo2020hfel, yang2020energy} in the literature, we do not consider the energy consumption of the base station (i.e., the central server).

\subsubsection{Transmission energy consumption}

Generally, transmission time consists of the uplink and downlink time. However, since the output power of the base station is significantly larger than a user device's maximum uplink transmission power, the downlink time is much shorter than the uplink time. 
Thus, we neglect the downlink time and only consider the uplink time, as in~\cite{dinh2020federated,yang2020energy,luo2020hfel}.

Following Shannon's pioneer work \cite{shannon1949communication}, we express the data transmission rate, $r_n$, as
\begin{align} \label{def:data_rate}
    \color{black}\textstyle{r_n = B_n\log_2 (1+ \frac{\mathbb{E}[\mathcal{G}_n]p_n}{N_0B_n})},
\end{align}
where $B_n$ is the bandwidth allocated to user $n$, $N_0$ is the power spectral density of Gaussian noise, $p_n$ is the transmission power. 
\textcolor{black}{Besides, we assume that at the $r$-th global round, the channel gain $g_{n,r}$ is a random variable obeying a distribution $\mathcal{G}_n$. $\mathbb{E}[\mathcal{G}_n]$ is the expectation of the channel gain. We will explain why we use $\mathbb{E}[\mathcal{G}_n]$ instead of $g_{n,r}$ in Section \ref{subsec:total_t_def}.}
The data transmission will happen in each global aggregation. 
If the transmission data size of user device $n$ is $d_n$ (bits), the uplink transmission time from user device $n$ to the base station is
\begin{align} \label{t_up_n}
    \textstyle{T^{trans}_n = d_n/r_n}.
\end{align}
The \textbf{transmission energy consumption} of user device $n$ is 
\begin{align} \label{e_trans_n}
    \textstyle{E^{trans}_n = p_nT^{trans}_n}.
\end{align}

\subsubsection{Local computation energy consumption}
Following \cite{mao2016dynamic}, and assuming that the CPU frequency $f_n$ { (i.e., the number of CPU cycles per second)} of device $n$ does not change once it is decided at the beginning of the FL process, we express the energy consumed in one \textbf{local} iteration of user device $n$ as: 
\begin{align}\label{equa:energy consump}
   \kappa c_n D_n f_n^2,
\end{align}
where $\kappa$ is the effective switched capacitance, and $c_n$ is the number of CPU cycles per \textit{standard sample}. A standard sample means a square video frame with a resolution of $\overline{\text{s}}_{\text{standard}}^2$ pixels.

However, each MAR device may not always use the exact video frame resolution to train its local model.
For instance, if a user forgets to charge the VR device, the device cannot use high-resolution video frames to train its model under such a low-battery condition. 
Intuitively, processing higher-resolution video frames requires more CPU cycles, while lower-resolution video frames need less.
Thus, the calculation of CPU cycles $c_nD_n$ in Eq. (\ref{equa:energy consump}) cannot stand for the cycles needed for high-resolution video frames nor low-resolution video frames.

We now analyze the needed CPU cycles for different MAR video frame resolutions. For simplicity, our paper considers only square video frames (i.e., images), which means that the height equals the width. On device $n$, let the video frame resolution be $s_n \times s_n$ pixels. Sometimes we abuse the description and refer to $s_n$ as the video frame resolution for simplicity.

We next focus on how to quantify the CPU cycles needed for video frames with different resolutions. For a video frame with a resolution of $s_n^2$ pixels and $s_n > \overline{\text{s}}_{\text{standard}}$, it  needs more CPU cycles to be processed than the standard sample. Similarly, for a video frame with a lower resolution of $s_n^2$ pixels and $s_n < \overline{\text{s}}_{\text{standard}}$, it will need fewer CPU cycles than the standard sample.

\begin{figure*}[htbp]
    \centering
    \includegraphics[width=11cm]{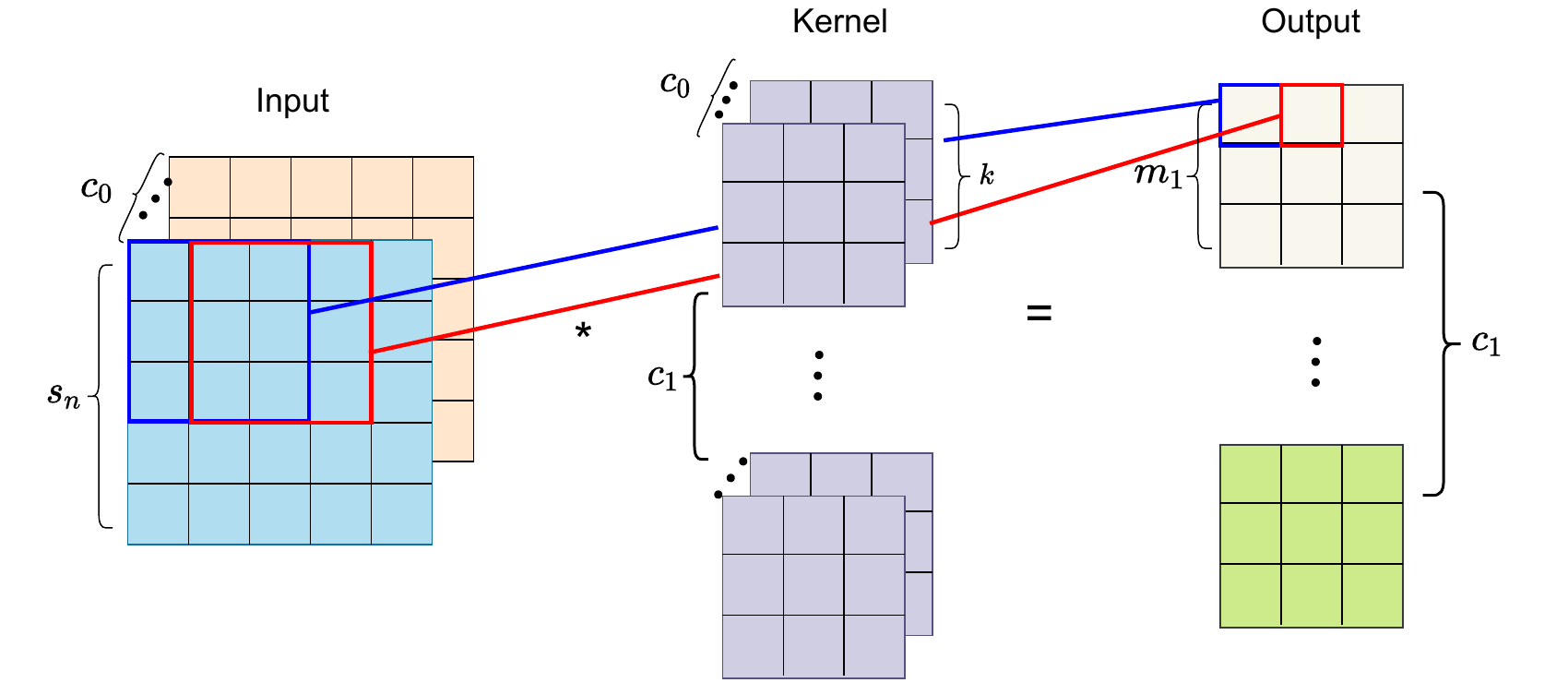}
    \caption{Illustration of the 1st convolutional layer in a CNN. In this example, \texttt{padding} is 0 and \texttt{stride} is 1. $c_0$ is the default number of channels (e.g., in a color image, we have three channels). $c_1$ is the number of kernels, the same as the number of output channels. $m_1$ is calculated by Eq.~(\ref{equa:output_feature_map}). By applying the kernel to the input image, doing element-wise multiplication and adding each element weighted by the kernel, we get the output by $c_0k^2m_1^2$ calculations. Since we have $c_1$ kernels, the total number of calculations is $c_0k^2c_1m_1^2$. If we have $L$ convolutional layers, we can get the time complexity in (\ref{equa:cnn_complexity}). \vspace{-10pt}}
    \label{fig:cnn}
\end{figure*}

To perform object detection, our metaverse system incorporates You Look Only Once (YOLO) algorithm \cite{redmon2016you}, whose main structure consists of convolutional neural networks (CNNs) \cite{krizhevsky2012imagenet}. From~\cite{he2015convolutional}, the time complexity of a CNN can be given as 
\begin{align} \label{equa:cnn_complexity}
    \textstyle{\mathcal{O}(\sum_{l=1}^{L}c_{l-1} \cdot k^2 \cdot c_l \cdot m_l^2)},
\end{align}
where $L$ is the total number of convolutional layers; $k$ refers to the kernel length; for $l \in \{1, 2, \ldots, L\}$, $c_{l-1}$ is the number of input channels to the $l$-th layer, $c_l$ is the number of kernels (i.e., the number of output channels after the $l$-th layer), and $m_l$ means the output feature map's spatial size. From~\cite{he2015convolutional}, in the first layer, $m_1$ can be computed by
\begin{align} \label{equa:output_feature_map}
    m_1 = (s_n - k + 2 * \texttt{padding})/\texttt{stride} + 1.
\end{align}
In the above equation, \texttt{padding} and \texttt{stride} are set in a CNN~\cite{he2015convolutional}, and $s_n \times s_n$ is the resolution of the input image on MAR device $n$.
In CNN, the current layer's output feature map is the next layer's input.
Thus, the input image size affects the spatial size of output feature maps of subsequent layers. 
Fig. \ref{fig:cnn} is a simple illustration of the first convolutional layer in a CNN to explain how we get the time complexity in (\ref{equa:cnn_complexity}). 

Because we want to find the relationship between the CPU cycles and input images, we only focus on how the input images affect the time complexity. Therefore, from Eq. (\ref{equa:cnn_complexity}) and (\ref{equa:output_feature_map}), we find the time complexity with respect to the input image resolution $s_n$ is $\mathcal{O}(s_n^2)$. Recall that $c_n$ is the number of CPU cycles on device $n$ per \textit{standard sample} with a resolution of $\overline{\text{s}}_{\text{standard}} \times \overline{\text{s}}_{\text{standard}}$ pixels. Hence, for simplicity, we just consider that the number of CPU cycles on device $n$ to process a video frame with a resolution of $s_n\times s_n$ pixels is given by $\zeta s_n^2 \times c_n $ for $\zeta := \frac{1}{\overline{\text{s}}_{\text{standard}}^2}$. Then the required CPU cycles to process $D_n$ video frames on device $n$ is
\begin{align}
     \textstyle{\zeta s_n^2 c_n D_n},
\end{align}
so the \textbf{local computation energy consumption in $R_l$ local iterations for one global round} is
\begin{align} \label{e_cmp_n}
   \textstyle{ E_n^{cmp} = \kappa R_l \zeta s_n^2 c_nD_n f_n^2},
\end{align}
where $R_l$ is the number of local iterations between two global aggregations. Let $R_g$ be the number of global rounds.
The total energy consumption is 
\begin{align} \label{equa:total_energ}
\textstyle{\mathcal{E} = R_g\sum_{n=1}^{N} (E_n^{trans} + E_n^{cmp})}. \end{align}

\subsection{Total Completion Time} \label{subsec:total_t_def}
The total completion time, $\mathcal{T}$, of the FL system model over the wireless network comprises local computation time and data transmission time. The local computation time of device $n$ in one \textbf{global} iteration, $T^{cmp}_n$, is 
\begin{align} \label{t_cmp_n}
    \textstyle{T_n^{cmp} = R_l\frac{\zeta s_n^2c_nD_n}{f_n}},
\end{align}
where $R_l$ is the number of local iterations on each device $n$, $\zeta s_n^2 c_nD_n$ is the number of  CPU cycles, and $f_n$ refers to the CPU frequency (i.e., the number of  CPU cycles per second).

Then, with the transmission time given by (\ref{t_up_n}),
the total completion time becomes 
\begin{align}
\textstyle{\mathcal{T}= R_g\max_{n \in \mathcal{N}} \{T_n^{cmp} + T_n^{trans}\}}. \label{jzEqT}
\end{align} 
As we have mentioned, since the base station's output power is larger than the maximum uplink transmission power of a user device, the downlink time is much smaller than the uplink time. Hence, we ignore the downlink time, as in other federated learning work~\cite{luo2020hfel,yang2020energy,dinh2020federated}.

\textcolor{black}{Besides, remember that we use $\mathbb{E}[\mathcal{G}_n]$ as the expectation of the channel gain. 
Since $T_n^{cmp}+T_n^{trans}\leq \max_{n\in\mathcal{N}}\{ T_n^{cmp}+T_n^{trans}\}$, 
$\mathbb{E}[T_n^{cmp}+T_n^{trans}]=\mathbb{E}[T_n^{cmp}]+\mathbb{E}[T_n^{trans}]$ $\leq$ $\mathbb{E}[\max_{n\in\mathcal{N}}\{ T_n^{cmp}+T_n^{trans}\}]$. 
Denote $T_n^{trans}(g_{n,r})$ as the transmission time with respect to $g_{n,r}$.
Note that according to Jensen's inequality and $T_n^{trans}(g_{n,r})$ is convex, the transmission time $T_n^{trans}(\mathbb{E}[g_{n,r}])$ (i.e., $T_n^{trans}(\mathbb{E}[\mathcal{G}_n])$) is the lower bound of $\mathbb{E}[T_n^{trans}(g_{n,r})]$. 
Then, $T_n^{trans}(\mathbb{E}[\mathcal{G}_n])+\mathbb{E}[T_n^{cmp}]\leq \mathbb{E}[\max_{n\in\mathcal{N}}\{ T_n^{cmp}+T_n^{trans}\}].$
Therefore, replacing the channel gain $g_{n,r}$ by the expectation $\mathbb{E}(\mathcal{G}_n)$ can obtain the lower bound of transmission time. In this paper, we aim at minimizing the lower bound of total energy and time consumption.
}

\subsection{Accuracy Model}  \label{sec:Accuracy}
An accuracy metric $\mathcal{A}$ is considered in the optimization objective. 
Since the optimization variables include the video frame resolution $s_n$ and if our goal is to just minimize the weighted sum of total energy consumption $\mathcal{E}$ and total delay $\mathcal{T}$, less $s_n$ is always better.
Hence, we incorporate $\mathcal{A}$ into the objective function to make the optimization of $s_n$ less trivial. 

Let $\mathcal{A}$ be the final training accuracy of the whole federated learning process and consider it as a function of $s_1, s_2, \ldots, s_N$. This can be written as
 $\mathcal{A}(s_1, s_2, \ldots, s_N)$ for simplicity.

\textcolor{black}{Based on the result from~\cite{liu2018edge}, we assume that $\mathcal{A}(s_1, s_2, \ldots, s_N)=\sum_n^N A_n(s_n)$ is concave (reasons for enforcing the accuracy metric $\mathcal{A}=\sum_n^N A_n(s_n)$ are explained in Section \ref{sec:prob_formu}).}

\section{Optimization Problem Formulation} \label{sec:prob_formu}

This section formulates the optimization problem. We obtain the formulation by discussing the optimization objective function, optimization variables, and constraints, respectively.

\vspace{-10pt}
\subsection{Optimization Problem} \label{sec:prob_Optimization}

For our federated learning with augmented reality, the optimization problem is as follows (the meanings of the symbols will be explained below):
\begin{subequations} \label{equa:min}
\begin{align}
\min_{ \{ B_n,p_n,f_n,s_n \}_{n\in \mathcal{N}} } &~ \textstyle{w_1 \mathcal{E} +w_2 \mathcal{T} -\rho \mathcal{A}}, \label{jzOPT-objective} \tag{\ref{equa:min}} \\[-5pt]
   \text{subject to:~} 
   & \textstyle{\sum_{n=1}^N B_n \le B,~B_n \ge 0,~\forall n \in \mathcal{N}}, \label{constra:bandwidth}\\
   & \textstyle{p_n^{min} \le p_n \le p_n^{max},~\forall n \in \mathcal{N}}, \label{pminpmax} \\[-5pt]
   & \textstyle{f_n^{min} \le f_n \le f_n^{max},~\forall n \in \mathcal{N}}, \label{fminfmax} \\[-5pt]
   & \textstyle{{\color{black}s_n \in \{\overline{\text{s}}_{1}, \cdots, \overline{\text{s}}_{M}\},~\forall n \in \mathcal{N}}}, \label{sminsmax}
\end{align}
\end{subequations}
where constraint (\ref{constra:bandwidth}) defines the upper limit of total bandwidth,  (\ref{pminpmax}) sets the range of each device's transmission power,  (\ref{fminfmax}) is the range of CPU frequency of each device, and  \textcolor{black}{(\ref{sminsmax}) gives $M$ choices for the video frame resolution: $\overline{\text{s}}_{1}, \cdots, \overline{\text{s}}_{M}$ which are constants (the overlined texts and normalized fonts are used to differentiate the constants $\overline{\text{s}}_{m}|_{m=1,\ldots,M}$  from the variables $s_n|_{n=1,\ldots,N}$) and ranked in the ascending order, i.e., $\overline{\text{s}}_{1}$ is the minimum resolution and $\overline{\text{s}}_{M}$ is the maximum resolution. }
In the objective function~(\ref{equa:min}),
the total energy consumption $\mathcal{E}$ is given in Eq.~(\ref{equa:total_energ}), and the total delay $\mathcal{T}$ is given in Eq.~(\ref{jzEqT}). The weight parameters $w_1$, $w_2$, and $\rho$ are all \mbox{non-negative}, and we enforce $w_1 + w_2 = 1$ without loss of generality.
 For devices in a low-battery condition, we can have $w_1 = 1$ and $w_2 = 0$. In contrast, if the  system is delay-sensitive, we can choose $w_1 = 0$ and $w_2 = 1$. $\rho$ is the weight parameter of the model accuracy. Larger (resp., smaller) $\rho$ means that the system puts more (resp., less) emphasis on the accuracy part.

\textbf{Reasons for enforcing the accuracy metric $\mathcal{A}$ as $\sum_{n=1}^{N} A_n(s_n)$ and constraining $s_n$ via (\ref{sminsmax}).} The optimization for other variables $B_n,p_n,f_n$ is already quite complex, as shown in our recent conference version~\cite{ICDCS}. A complex expression for $\mathcal{A}$ or more value choices for $s_n$ can cause the overall optimization to be intractable.
Besides, our simple consideration is the first step. Future directions include considering more complex $\mathcal{A}$ and more value choices $s_n$.

Before solving the optimization problem, we now transform the problem into a form that is easier to handle.
  From Eq.~(\ref{jzEqT}), $\mathcal{T} $ equals $R_g\max_{n \in \mathcal{N}} \{T_n^{cmp} + T_n^{trans}\}$. Such max function in the above expression of $\mathcal{T} $ makes the minimization in~(\ref{equa:min}) hard to solve. To circumvent this difficulty, we introduce an auxiliary variable $T$, replace $\mathcal{T} $ by $R_g T$, and add an extra constraint $T_n^{cmp} + T_n^{trans} \leq T$ for $n\in \mathcal{N}$. 
  Then, the problem writes as follows:
\begin{subequations} \label{equa:min2}
\begin{align}
\min_{ \{ B_n,p_n,f_n,s_n \}_{n\in \mathcal{N}} , T} &~ w_1R_g \sum_{n=1}^{N} (p_nT_n^{trans}+\kappa R_l\zeta s_n^2c_nD_nf_n^2)\notag\\[-5pt]
&+ w_2 R_gT -\rho\sum_{n=1}^{N}A_n(s_n) \tag{\ref{equa:min2}} \\
    \text{subject to:~} & \text{(\ref{constra:bandwidth}),~(\ref{pminpmax}),~(\ref{fminfmax}),~(\ref{sminsmax}),}\nonumber\\[-5pt]
   & T_n^{cmp} + T_n^{trans} \leq T, ~\forall n\in \mathcal{N}.\label{constra:time}  
\end{align}
\end{subequations}

\textbf{Difficulty of solving problem (\ref{equa:min2})}. Since the formulation of $T_n^{trans}$ and some constraints are non-convex with respect to the optimization variables, problem (\ref{equa:min2}) is \mbox{non-convex}, which is intractable. Thus, we must explore other ways to transform this problem to make it solvable.

\subsection{Problem Decomposition}
To continuously transform the problem (\ref{equa:min2}) into a simpler form, we thereby separate problem (\ref{equa:min2}) into two subproblems. One has $p_n$ and $B_n$ as optimization variables since $p_n$ and $B_n$ are in the same term, and the other has $f_n$, $s_n$ and $T$ as optimization variables.

Two minimization problems are as follows:
{
\begin{align}
\text{\textbf{Subproblem 1 (SP1):}}\\~\min_{ \{f_n, s_n \}_{n\in \mathcal{N}}, T} ~~~~~~~~~~~&\hspace{-40pt}~ w_1 R_g\sum_{n=1}^{N} \kappa R_l\zeta s_n^2 c_nD_nf_n^2+ w_2R_g T \notag\\[-5pt]
&-\rho\sum_{n=1}^N A_n(s_n)  \label{equa:min5}  \\[-5pt]
    \text{subject to:} &  ~\text{(\ref{fminfmax}),}~ \text{(\ref{sminsmax}),}~\text{(\ref{constra:time}).}  \notag 
    \end{align}
    \begin{align}
    \text{\textbf{Subproblem 2 (SP2):}}\\~\min_{\{p_n, B_n\}_{n\in \mathcal{N}}} &~ w_1R_g \sum_{n=1}^{N} p_n\frac{d_n}{B_n \log_2(1+\frac{p_ng_n}{N_0B_n})} \label{equa:min6} \\[-5pt]
    \text{subject to:} &~ \text{(\ref{pminpmax}),~(\ref{constra:bandwidth}),}~\text{(\ref{constra:time}).} \notag
\end{align}

}

In the rest of the paper, for a vector $\bm{v}$, its $n$-th dimension is denoted as $v_n$.
Based on the above notation system, we have 
$\bm{p}:=[p_1, p_2, \ldots, p_N]$, $\bm{B}:=[B_1, B_2, \ldots, B_N]$, {$\bm{f}:=[f_1, f_2, \ldots, f_N]$},  and $\bm{s}:=[s_1, s_2, \ldots, s_N]$.
Our algorithm to solve the problem (\ref{equa:min2}) is based on the Block coordinate descent (BCD) technique. BCD optimizes a group of optimization variables (or blocks of indices) and fixes other blocks at each iteration \cite{yu2019enabling}. We randomly choose feasible $(\bm{p}, \bm{B})$ at first, and solve Subproblem 1 to get the optimal ($\bm{f}$, $\bm{s}$) (note that $T$ is an auxiliary variable). Then, we solve Subproblem 2 to get the optimal $(\bm{p}, \bm{B})$. The whole process consists of iteratively solving Subproblems 1 and 2 until convergence.

\section{Solution of The joint optimization problem} \label{sec:solution_joint_prob}
In this section, we will explain how to solve Subproblems 1 and 2.

\subsection{Solution to Subproblem 1 } \label{secSubproblem 1}
Since the video frame resolution $s_n$ is a discrete variable, we relax it into a continuous variable $\hat{s_n}$ to make Subproblem 1 easier to solve. 
Then, Subproblem 1 becomes
\begin{subequations} \label{equa:min7}
\begin{align} 
    \textbf{\textit{SP1\_v1:}}&\min_{\{f_n, \hat{s_n}\}_{n\in \mathcal{N}}, T} w_1 R_g\sum_{n=1}^{N} \kappa R_l\zeta \hat{s_n}^2 c_nD_nf_n^2 \notag\\
    &+ w_2R_g T - \rho\sum_{n=1}^N A_n(\hat{s_n}) \tag{\ref{equa:min7}} \\[-5pt]
    \text{subject to:} &  ~\text{(\ref{fminfmax}),}~\text{(\ref{constra:time}),} ~ \overline{\text{s}}_{1} \le \hat{s_n} \le \overline{\text{s}}_{M},~\forall n\in \mathcal{N}.
\end{align}
\end{subequations}
It is easy to verify that the objective function and constraints of variables $f_n$, $\hat{s_n}$ and $T$ in (\ref{equa:min7}) are convex. Then, we treat the optimization problem by the Karush--Kuhn--Tucker (KKT) approach.

Since applying KKT conditions is a common approach in convex optimization, we place the derivation in Appendix \ref{sol_to_sp1}. In short, we first introduce the Lagrange function of \textbf{\textit{SP1\_v1}} and define the Lagrange multiplier $\bm{\lambda}=[\lambda_1,\cdots,\lambda_N]$. After applying KKT conditions to the Lagrange function, we derive the expression of $\bm{f}$ in $\bm{\lambda}$ and thus derive the dual problem of \textcolor{black}{\textbf{\textit{SP1\_v1}}}. \textcolor{black}{In this paper, we could use CVX \cite{grant2014cvx} 
to solve the problem because of its convexity. In Appendix \ref{sol_to_sp1}, we discuss the special case that $A_n(\hat{s}_n)$ is linear and find its dual problem to reduce the time complexity.} Then, we are able to calculate $\bm{f}$ and $\bm{\hat{s}}$ through Eq. (\ref{relation_f_s_lamda}) (in Appendix \ref{sol_to_sp1}).

Due to $f_{n}^{\min} \le f_n \le f_n^{\max}$, $f_n$ is set as follows with $f_n^*$ in Eq. (\ref{relation_f_s_lamda}):
\begin{align}
    f_n  & = \min(f_n^{\max},\max(f_n^*,~f_n^{\min})).\label{relation_f_lamda2}  
\end{align}
Since the video frame resolution $s_n$ is discrete, we rank all resolution options in the ascending order as $\{\overline{\text{s}}_1, \cdots, \overline{\text{s}}_M\}$ and map $\hat{s_n}$ to $s_n$ according to the formula below  \textcolor{black}{
\begin{align} 
    s_n\!=\! 
    \begin{cases} 
    & \overline{\text{s}}_{M}, ~\text{if}~\hat{s_n} \geq \frac{\overline{\text{s}}_{M}+\overline{\text{s}}_{M-1}}{2},\\
   & \overline{\text{s}}_{M-1}, ~\text{if}~ \frac{\overline{\text{s}}_{M-2}+\overline{\text{s}}_{M-1}}{2} \!\le\! \hat{s_n} \!<\! \frac{\overline{\text{s}}_{M-1}+\overline{\text{s}}_{M}}{2},\\
    &\cdots,\\
    & \overline{\text{s}}_{1}, ~\text{if}~\hat{s_n} < \frac{\overline{\text{s}}_{1}+\overline{\text{s}}_{2}}{2},
    \end{cases} 
    \label{proposedmap} 
\end{align}
}
where $\hat{s_n}$ is in Eq. (\ref{relation_f_s_lamda}).
There may exist other methods to map $\hat{s_n}$ to $s_n$, which may be more involved and require more computations. Our proposed mapping in Eq.~(\ref{proposedmap}) above is simple without complex computations. The effectiveness of our mapping is further confirmed by experimental results in Section~\ref{sec:experiments}.

\subsection{Subproblem 2}

Subproblem 2 is a typical sum-of-ratios minimization problem, which is NP-complete, and thus very hard to solve~\cite{jong2012efficient}. Firstly, we define
$G_n(p_n, B_n) := B_n\log_2(1+\frac{p_ng_n}{N_0B_n})$
to simplify the notations.
Additionally, (\ref{constra:time}) can be written as $r_n \ge r_n^{\min}$, where $r_n^{\min} = \frac{d_n}{T-\frac{R_lc_nD_n}{f_n}}$.
In fact, $G_n(p_n, B_n)$ is the same as the data rate $r_n$ defined in (\ref{def:data_rate}).

Then, we transform Subproblem 2 into the epigraph form through adding an auxiliary variable $\beta_n$ and let $\frac{p_nd_n}{G_n(p_n, B_n)} \le \beta_n$. Then, we get an equivalent problem called \textbf{\textit{SP2\_v1}}.
\begin{subequations} \label{equa:subproblem2_transform}
\begin{align}
    \textbf{\textit{SP2\_v1:}}&~\textstyle{\min_{p_n, B_n, \beta_n}  w_1R_g \sum_{n=1}^{N} \beta_n }\tag{\ref{equa:subproblem2_transform}}\\
    \text{subject to~} &\text{(\ref{constra:bandwidth})},~\text{(\ref{pminpmax})},~ \notag \\
    & \textstyle{G_n(p_n, B_n) \ge r_n^{\min}, ~\forall n \in \mathcal{N}}, \label{constra:subproblem2_data_rate_min}\\
    & \textstyle{p_nd_n -\beta_nG_n(p_n, B_n) \le 0,~\forall n \in \mathcal{N}}, \label{constra:epi_data_rate} 
\end{align}
\end{subequations}
where $\beta_n$ is an auxiliary variable. Because $\frac{p_nd_n}{G_n(p_n, B_n)} \le \beta_n$ and $G_n(p_n, B_n) > 0$, we can get constraint (\ref{constra:epi_data_rate}).
However, due to constraint (\ref{constra:epi_data_rate}), problem \textbf{\textit{SP2\_v1}} is \mbox{\mbox{non-convex}}, which is still intractable. To continue transforming this problem, we need to give the following lemma.

\begin{lemma}[Proved in Appendix \ref{appen:proof_lemma_requis}] \label{lemma:requisite_fra_prog}
$G_n(p_n, B_n)$ is a concave function.
\end{lemma}

The original objective function in Subproblem 2 is the sum of fractional functions. It is obvious that $p_nd_n$ is convex, and Lemma \ref{lemma:requisite_fra_prog} reveals $G_n(p_n, B_n)$ is concave. With such a trait, problem \textbf{\textit{SP2\_v1}} can be transformed into a subtractive-form problem via the following theorem.

\begin{theorem} \label{theor:equal_prob_subp2}

If $(\bm{p}^*, \bm{B}^*, \bm{\beta}^*)$ is the solution of problem \textbf{SP2\_v1}, there exists $\bm{\nu}^*$ satisfying that $(\bm{p}^*, \bm{B}^*)$ is a solution of the following problem with $\bm{\nu} = \bm{\nu}^*$ and $\bm{\beta} = \bm{\beta}^*$.
\begin{subequations} \label{equa:subproblem2_lagrangian_min}
\begin{align}
\textbf{SP2\_v2:}&~ \textstyle{ \min_{p_n, B_n} \sum_{n=1}^N \nu_n(p_nd_n-\beta_nG_n(p_n, B_n))} \tag{\ref{equa:subproblem2_lagrangian_min}} \\[-5pt]
 \text{subject to:}&~\text{(\ref{constra:bandwidth})},~\text{(\ref{pminpmax})},~\text{(\ref{constra:subproblem2_data_rate_min}).} \notag
\end{align}
\end{subequations}

Moreover, when $\bm{\nu} = \bm{\nu}^*$ and $\bm{\beta} = \bm{\beta}^*$, ($\bm{p}^*, \bm{B}^*$) satisfies the following equations.
\begin{align}
 &\textstyle{   \nu_n^* = \frac{w_1R_g}{G_n(p_n^*, B_n^*)}}, \text{ and }
    \textstyle{\beta_n^* = \frac{p_n^*d_n}{G_n(p_n^*, B_n^*)}}, \text{ for} ~n=1,\cdots, N. \label{equa:beta_update}
\end{align}
\end{theorem}
Proof: The proof follows directly from Lemma 2.1 of \cite{jong2012efficient}.

\subsection{Solution to Subproblem 2}
\textbf{Theorem \ref{theor:equal_prob_subp2}} indicates that problem \textbf{\textit{SP2\_v2}} has the same optimal solution $(\bm{p^*, \bm{B}^*})$ with problem \textbf{\textit{SP2\_v1}}. Consequently, we can first provide $(\bm{\nu}, \bm{\beta})$ to obtain a solution $(p_n, B_n)$ through solving problem \textbf{\textit{SP2\_v2}}. Then, the second step is to calculate $(\bm{\nu}, \bm{\beta})$ according to (\ref{equa:beta_update}). The first step can be seen as an inner loop to solve problem \textbf{\textit{SP2\_v2}} and the second step is the outer loop to obtain the optimal $( \bm{\nu},\bm{\beta})$. The iterative optimization   is given in Algorithm 1, which is a Newton-like method. The original algorithm is provided in \cite{jong2012efficient}. In Algorithm 1, we define 
\begin{align}
     \phi_1(\bm{\beta}) &\!=\! [-p_nd_n+\beta_nG_n(p_n, B_n)|_{n = 1,\cdots, N}]^T \!,\\
     \phi_2(\bm{\nu})& \!=\! [-w_1R_g+ \nu_nG_n(p_n, B_n)|_{n = 1,\cdots, N}]^T,
\end{align}
which are two vectors in $\mathbb{R}^{N}$, and
    $\phi(\bm{\beta}, \bm{\nu}) = [\phi_1^T(\bm{\beta}), \phi_2^T(\bm{\nu})]^T$,
which is a vector in $\mathbb{R}^{2N}$. Additionally, the Jacobian matrices of $\phi_1(\bm{\beta})$ and $\phi_2(\bm{\nu})$ are
\begin{align}
   \phi_1^{\prime}(\bm{\beta})&=diag(G_n(p_n, B_n)|_{n = 1,\cdots, N}),\\
   \phi_2^{\prime}(\bm{\nu})&=diag(G_n(p_n, B_n)|_{n = 1,\cdots, N}),
\end{align}
where $diag()$ represents a diagonal matrix. \cite{jong2012efficient} proves that when $\phi(\bm{\beta}, \bm{\nu})=\bm{0}$, the optimal solution $(\bm{\nu }, \bm{\beta})$ is achieved. Simultaneously, it satisfies (\ref{equa:beta_update}).
Thus, a Newton-like method can be used to update $(\bm{\nu }, \bm{\beta})$, which is done in (\ref{update_nu_beta}).
\begin{algorithm}
\label{algo:MN}
\caption{Optimization of $(\bm{p}, \bm{B})$ and $(\bm{\nu}, \bm{\beta})$}
Initialize $i = 0$, $\xi \in (0, 1)$, $\epsilon \in (0, 1)$. Given feasible ($\bm{p}^{(0)}$, $\bm{B}^{(0)}$). \\
Calculate 
$\nu_n^{(i)} = \frac{w_1R_g}{G_n(p_n^{(i)}, B_n^{(i)})},~\beta_n^{(i)} = \frac{p_nd_n}{G_n(p_n^{(i)}, B_n^{(i)})}$
for $n=1,\cdots, N$.\\
\Repeat{$\phi(\bm{\beta}^{(i)},\bm{\nu}^{(i})) = \bm{0}$ or the number of iterations reaches the maximum iterations $i_0$}{
Obtain $(\bm{p}^{(i+1)}$, $\bm{B}^{(i+1)})$ through solving \textbf{\textit{SP2\_v2}} according to \textbf{Theorem \ref{theor:express_B_p}} by using CVX given $(\bm{\nu}^{(i)}, \bm{\beta}^{(i)})$.

Let $j$ be the smallest integer which satisfies
\begin{align} \label{newton_method}
    | \phi(\bm{\beta}^{(i)}+\xi^j\bm{\sigma_1}^{(i)}, \bm{\nu}^{(i)}+\xi^j\bm{\sigma_2}^{(i)}) |  \le (1-\epsilon\xi^j) | \phi(\bm{\beta}^{(i)}, \bm{\nu}^{(i)}) | ,
\end{align}
where 
\begin{align}
    \bm{\sigma_1}^{(i)} = -[\phi_1^{\prime}(\bm{\beta}^{(i)})]^{-1}\phi_1(\bm{\beta}^{(i)}),\notag\\[-5pt] \bm{\sigma_2}^{(i)} = -[\phi_2^{\prime}(\bm{\nu}^{(i)})]^{-1}\phi_2(\bm{\nu}^{(i)}).
\end{align}

Update 
\begin{align} \label{update_nu_beta}
    (\bm{\beta}^{(i+1)},\bm{\nu}^{(i+1)}) =( \bm{\beta}^{(i)} + \xi^j\bm{\sigma_1}^{(i)},\bm{\nu}^{(i)}+\xi^{j}\bm{\sigma_2}^{(i)}).
\end{align}

Let $i \leftarrow i+1$.
}
\end{algorithm}

Until now, the remaining problem needed to focus on is problem \textbf{\textit{SP2\_v2}}. Remember $\bm{\nu}$ and $\bm{\beta}$ are fixed in this problem, which are seen as constants. Given Lemma \ref{lemma:requisite_fra_prog}, $p_nd_n$ is convex, and three constraints \text{(\ref{pminpmax})}~\text{(\ref{constra:bandwidth})}~\text{(\ref{constra:subproblem2_data_rate_min})} are convex, the objective function (\ref{equa:subproblem2_lagrangian_min}) in problem \textbf{\textit{SP2\_v2}} is convex. Thus, KKT conditions are sufficient and necessary optimality conditions for finding the optimal solution.
Since the process of applying KKT conditions to \textbf{\textit{SP2\_v2}} is similar to that of \textbf{\textit{SP1\_v1}} in (\ref{equa:min7}), we skip the derivation here and put it in Appendix \ref{appen:proof_express_bp}.

Then, \textbf{Theorem \ref{theor:express_B_p}} is given to find the optimal bandwidth and transmission power for each device. 
Note that $\tau_n|_{n=1,\ldots,N}$ and $\mu$ are non-negative Lagrange multipliers and the Lagrange function of \textbf{\textit{SP2\_v2}} is written in Appendix \ref{appen:proof_express_bp}.

\begin{theorem} \label{theor:express_B_p}
The optimal bandwidth $\bm{B}$ and transmission power $\bm{p}$ are expressed as 
\begin{align}
\hspace{-2pt} & B_n^* = 
   \begin{cases} 
   \hspace{-3pt}  \frac{r_n^{\min}}{\log_2(1+\Lambda_n)},~\text{if}~\tau_n \neq 0, 
   \text{Solution~to~problem~(\ref{SP2_v3}),}~ \text{if}~\tau_n = 0,
   \end{cases} \nonumber 
   \\
 & \text{and }   p_n^* = \min(p_n^{\max},\max(\Gamma(B_n),~p_n^{\min})),~\text{when}~\mu \neq 0,~ \nonumber 
\end{align}
where $\Lambda_n = \frac{(\nu_n\beta_n+\tau_n)g_n}{N_0d_n\nu_n \ln2}$ and $\Gamma(B_n) = (\frac{(\nu_n\beta_n+\tau_n)g_n}{N_0d_n\nu_n\ln2}-1)\frac{N_0B_n}{g_n}.$
\end{theorem}

We prove Theorem~\ref{theor:express_B_p} in Appendix \ref{appen:proof_express_bp}.
Then problem \textbf{\textit{SP2\_v2}} is solved.

\subsection{Resource Allocation Algorithm} \label{sec6}

Here we give the complete resource allocation algorithm based on BCD in Algorithm 2.

\begin{algorithm} \label{algo:resource_allocation_algorithm}
\caption{Resource Allocation Algorithm of FL-MAR system for the Metaverse.}
Initialize $ sol^{(0)} = (\bm{p}^{(0)}, \bm{B}^{(0)}, \bm{f}^{(0)}, \bm{s}^{(0)})$ of problem (\ref{equa:min2}), iteration number $k=1$.\\
\Repeat{$|sol^{(k)}-sol^{(k-1)}| \leq \epsilon_0$ or the number of iterations achieves $K$}{
Solve Subproblem 1 through solving problem (\ref{equa:subproblem1_dual}) with CVX given $(\bm{p}^{(k-1)}, \bm{B}^{(k-1)})$. Obtain $(\bm{f}^{(k)}, \bm{s}^{(k)})$ according to Eq. (\ref{relation_f_lamda2}) and Eq. (\ref{proposedmap}).\\
Solve Subproblem 2 through Algorithm 1 and obtain $(\bm{p}^{(k)}, \bm{B}^{(k)})$ with CVX.\\
$ sol^{(k)} = (\bm{p}^{(k)}, \bm{B}^{(k)}, \bm{f}^{(k)}, \bm{s}^{(k)})$.\\
Set $k \leftarrow k+1$.
}
\end{algorithm}
Algorithm 2 first initializes a feasible solution within the range of power $\bm{p}$ and bandwidth $\bm{B}$, and the sum of the bandwidth of each device cannot exceed $B$. 
Then, alternating optimization is applied. Solving problem (\ref{equa:subproblem1_dual}) given $(\bm{p}, \bm{B})$ makes sure we get ($\bm{f}$, $\bm{s}$, $T$) at this step. 
Subproblem~2 given ($\bm{f}$, $\bm{s}$, $T$) is an optimization problem focusing on $\bm{p}$ and $\bm{B}$.

\section{Convergence and Time Complexity} \label{sec:conv}
In this section, we analyze the convergence of our resource allocation algorithm (Algorithm~2) and present the time complexity of Algorithm 2.

\subsection{Convergence Analysis}
Note that Algorithm 2 iteratively solves Subproblem 1 to get ($\bm{f}, \bm{s}, T$) given ($\bm{p}, \bm{B}$), and Subproblem 2 to get ($\bm{p}, \bm{B}$) given ($\bm{f}, \bm{s}, T$). Then, we should analyze the convergence of Algorithm 1, which is to solve Subproblem 2.
In \cite{ICDCS}, we show that Algorithm 1 converges with a linear rate at any starting point in a solution set and a quadratic convergence rate of the solution's neighborhood.
Therefore, iteratively solving Subproblem 1 to get ($\bm{f}, \bm{s}$) and Subproblem 2 to get ($\bm{p}, \bm{B}$) will converge. 

\textcolor{black}{However, the convergence of our proposed algorithm does not guarantee the optimum. For SP1 given ($\bm{p}, \bm{B}$), we can guarantee the optimal values of $\bm{f}$, $\bm{s}$, and $T$ are computed (note that $\bm{s}$ is a discrete variable and the optimal $\bm{s}$ is only relative to the discrete values of $\bm{s}$). For SP2 given ($\bm{f}, \bm{s}, T$), we also find the globally optimal $(\bm{p}, \bm{B})$ \cite{jong2012efficient}. However, iteratively solving SP1 and SP2 via alternating optimization cannot guarantee the achievement of the optimal solution. Rather, it can only be theoretically demonstrated that the method will ultimately converge. As we stated in Section \ref{sec:prob_Optimization}, since the original problem is quite complex and non-convex, it would be intractable to find a method that can jointly optimize and find the global optimal solution.}

\subsection{Time Complexity}
For Algorithm 1, we use floating point operations (flops) to count the time complexity. One real addition/multiplication/division is counted as one flop \cite{he2013coordinated}. We mainly focus on analyzing Steps 4--6 because they take a major part of computational complexity. Step 4 utilizes CVX to solve problem \textit{\textbf{SP2\_v2}}. Besides, CVX invokes an interior-point algorithm to solve Semidefinite Programming (SDP) problems. Thus, given the solution accuracy $\epsilon_1 > 0$, the worst time complexity is $\mathcal{O}(N^{4.5}\log\frac{1}{\epsilon_1})$ \cite{luo2010semidefinite, zheng2021qos}. Additionally, 
Step 5 will take $\mathcal{O}((j+1)N)$ flops, where $j$ is the smallest integer satisfying (\ref{newton_method}). Our experiments show that typically $j$ is a small number. If $j=0$ (i.e., $\xi^j=1$), Step 6 is just the Newton method. Otherwise, it is a Newton-like method. Obviously, Step 6 takes $4N$ flops. Since the number of iterations is not larger than $i_0$, the time complexity of Algorithm 1 will not exceed $\mathcal{O}(i_0(N^{4.5}\log\frac{1}{\epsilon_1}+(j+1)N))$.

Algorithm 2 solves two subproblems successively by using the convex package CVX in each iteration. Steps 3 and 4 both use CVX to solve two subproblems. As we have mentioned above, the time complexity caused by CVX is $\mathcal{O}(N^{4.5}\log\frac{1}{\epsilon_1})$ given a solution accuracy $\epsilon_1>0$. Step 4 invokes Algorithm 1, and we have already analyzed the time complexity of Algorithm 1. Finally, the overall complexity of Algorithm 2 is $\mathcal{O}(K\cdot(i_0+1)N^{4.5}\log\frac{1}{\epsilon_1}+Ki_0\cdot(j+1)N)$.

\section{Experiments} \label{sec:experiments}

We conduct extensive experiments to illustrate the effectiveness of our proposed algorithm. We first present the parameter setting of our experiments in Section \ref{sec-exp-Parameter}, and then report various experimental results in other subsections. \textcolor{black}{To explore the influence of data heterogeneity on FL and MAR applications, we add experiments about image classification other than object detection by using MNIST and CIFAR-10 datasets.}

\subsection{Parameter Setting} \label{sec-exp-Parameter}
We describe the parameter settings of the experiments as follows. $N$ denotes the number of devices as 50. The devices are uniformly located in a circular area of size $500$ m $\times$ $500$ m, and the center is a base station. Between each device and the base station, the wireless channel's pass loss is modeled as \mbox{$128.1 + 37.6\log_{10}(\texttt{distance})$} along with the standard deviation of shadow fading, which is $8$ dB, where   \texttt{distance} is the distance from the device to the base station with kilometer as the unit. The power spectral density of Gaussian noise $N_0$ is $-174$ dBm/Hz.

Additionally, the number of local iterations $R_l$ is set as $10$, and the number of global aggregations $R_g$ is 100. 
The data size $d_n$ for each device to upload is set as $28.1$ kbits, based on the result of~\cite{yang2020energy}.  
Besides, each device has $D_n = 500$ samples. The number of CPU cycles per sample, $c_n$, is chosen randomly from $[1, 3]\times 10^4$.
The effective switched capacitance $\kappa$ is $10^{-28}$.
The maximum CPU frequency $f_n^{\max}$ and the maximum transmission power $p_n^{\max}$ for each device $n$ are $2$ GHz and $12$ dBm, respectively, so we just write them as $f^{\max}$ and $p^{\max}$ below. The minimum CPU frequency $f_n^{\min}$ is $0$ Hz, and the minimum transmission power $p_n^{\min}$ is \mbox{$0$ dBm.}  The total bandwidth $B$ is $20$ MHz.
\textcolor{black}{By default, the number of resolution options $M$ is set as $4$. Besides, $\overline{\text{s}}_{1} = 160$ pixels, $\overline{\text{s}}_{2} = 320$ pixels, $\overline{\text{s}}_{3} = 480$ pixels and $\overline{\text{s}}_{4} = 640$ pixels. In addition, $\overline{\text{s}}_{\text{standard}} = 160$ pixels.} Recall that the constant $\zeta$ equals $\frac{1}{\overline{\text{s}}_{\text{standard}}^2}$. \textcolor{black}{Although the accuracy function we use in the experiment is linear and the data is obtained from \cite{liu2018edge}, our algorithm is applicable to other concave functions.}
In all experiments, we  ``normalize''\footnote{The intuition for doing this is that since $\mathcal{E}$ and $\mathcal{T}$ can be understood as ``costs'' while $\mathcal{A}$ can be understood as ``gain'', then with $w_1 + w_2 = 1$, the combined coefficient $w_1 + w_2$ for the cost part is $1$, for better clarity.} the weight parameters such that $w_1 + w_2 = 1$; i.e., if $w_1 + w_2$ does not equal to $1$ at the beginning, we just divide all the parameters by $w_1+w_2$.  
Note that the above normalization can apply to all possible $(w_1, w_2, \rho)$ except for $(w_1=0, w_2=0, \rho=1)$ since it corresponds to the trivial problem of just maximizing the accuracy part $\mathcal{A}$.

\subsection{How does Our Optimization Algorithm Perform with respect to System Parameters?} \label{subsec:adj_weight_param}
In our optimization, we have three weight parameters $w_1$, $w_2$, and $\rho$, where $w_1 + w_2 = 1$ as explained above. If $w_1>w_2$, energy consumption will take a more dominant position in our optimization problem. If $w_2>w_1$, it is more important to optimize the delay instead of energy.
 Analogously, {if the weight parameter $\rho$ is larger, the model accuracy will take a more dominant position. To explore the effect of weight parameters of total energy and time consumption and model accuracy separately, we first show experiments under different $(w_1, w_2)$ with fixed $\rho$ and then show results under different $\rho$ with fixed $(w_1, w_2) = (0.5, 0.5)$. The algorithms to be compared with are summarized in Table \ref{tab:algorithm_comparison} (comparison with other work is in Section~\ref{secSuperiority}).
 
\begin{table*}[h]
\caption{Algorithm Comparison}
\centering \small
\begin{tabular}{|c|c|c|}
\hline
                     & \textbf{resource allocation ($f, p, B$)} & \textbf{video frame resolution $s_n$} \\ \hline
Proposed algorithm   & optimized                                & optimized                             \\ \hline
Benchmark (MinPixel) & random                                   & fixed                                 \\ \hline
RandPixel            & random                                   & random                                \\ \hline
\end{tabular}
\label{tab:algorithm_comparison} 
\end{table*}
 }

\subsubsection{Different ($w_1$, $w_2$) with fixed $\rho$}
In the experiments, we compare three pairs of $(w_1, w_2) = (0.9, 0.1)$, $(0.5, 0.5)$, and $(0.1, 0.9)$ with a benchmark algorithm under two scenarios. One is under different maximum transmission powers, and the other is under different maximum computational frequencies. 
$(w_1, w_2) = (0.9, 0.1)$ is to simulate devices under a low-battery condition, so the weight parameter of the energy consumption is set as 0.9. $(w_1, w_2) = (0.5, 0.5)$ represents the normal scenario, which means no preference to time or energy. Besides, $(w_1, w_2) = (0.1, 0.9)$ stands for the time-sensitive scenario. Thus, the algorithms in the comparison include:
\begin{itemize} 
    \item Under  weights $(w_1 = 0.9, w_2 = 0.1)$, $(w_1 = 0.5, w_2 = 0.5)$, or $(w_1 = 0.1, w_2 = 0.9)$.
    \item Benchmark algorithm (MinPixel): 1) When comparing energy consumption and completion time under different maximum transmission powers, for the $n$-th device, randomly select the CPU frequency $f_n$ from 0.1 to 2 GHz and set $p_n = p^{\max}$, $B_n = \frac{B}{N}$ and $s_n = \overline{\text{s}}_{1}$. 2) When comparing under different maximum CPU frequencies, randomly select the transmission power $p_n$ between 0 and 12 dBm and set $f_n = f^{\max}$, $B_n = \frac{B}{N}$ and $s_n = \overline{\text{s}}_{1}$.
\end{itemize}

When comparing energy consumption and completion time under different maximum transmission powers, $\rho=1$. When comparing under different maximum CPU frequencies, $\rho=10$.
To be more general, we run our algorithm at each pair of weight parameters 100 times and take the average value of energy and time consumption. At each time, 50 users are generated randomly.

\begin{figure*}[t!]
    \begin{adjustbox}{center,max width=1.1\textwidth}
    \begin{minipage}{.5\textwidth}
    \includegraphics[width=1\linewidth]{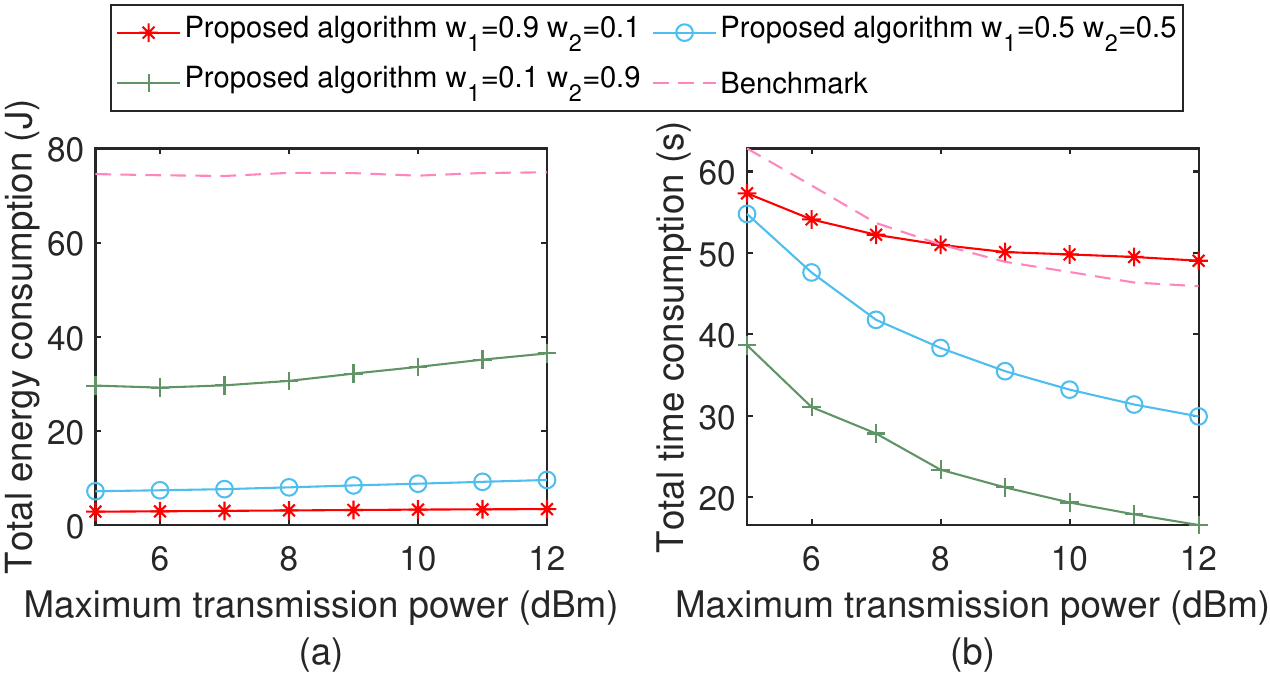}
    \caption{Total energy/time consumption with different weight parameters under different maximum transmit power limits. Here $\rho=1$.}
    \label{fig:total_e_t_different_p}
    \end{minipage}
    \begin{minipage}{.51\textwidth}
    \includegraphics[width=1\linewidth]{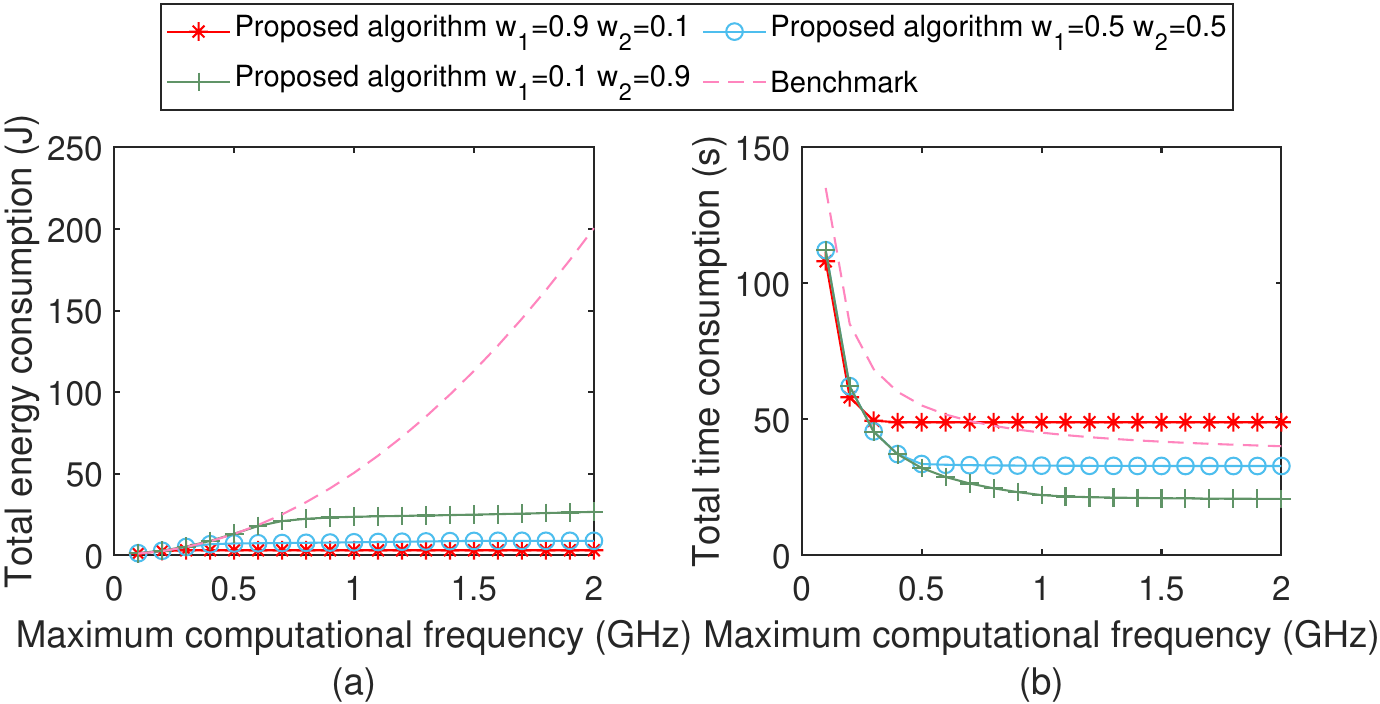}
    \caption{Total energy/time consumption with different weight parameters under different maximum CPU frequencies.\\ \textcolor{black}{Here $\rho=1$.}}
    \label{fig:total_e_t_different_f}
    \end{minipage}
    \end{adjustbox}
\end{figure*}

Fig. \ref{fig:total_e_t_different_p} shows the results of total energy consumption and total time consumption with three pairs of $(w_1, w_2)$ under different maximum transmit power limits, respectively. It can be observed when $w_1$ becomes larger and $w_2$ decreases, the total energy consumption becomes smaller, and the total time becomes larger. The reason is that if $w_1$ (resp., $w_2$) increases, our optimization algorithm will 
emphasize minimizing the energy cost (resp., time consumption). 

Additionally, in Fig. \ref{fig:total_e_t_different_p}(a), 
the green, black and red lines are all below the benchmark line with a wide gap. Note that the benchmark algorithm sets all video frame resolutions as $\overline{\text{s}}_{1}$, which means it saves much computational energy cost while sacrificing object detection accuracy. Even in this situation, our algorithm obtains better total energy consumption than the benchmark even at $w_1 = 0.1$ (this means the algorithm mainly focuses on minimizing the total time consumption). 

Correspondingly, Fig.~\ref{fig:total_e_t_different_p}(b) reveals that our time optimization is still better than the benchmark except for $(w_1=0.9,w_2 = 0.1)$ after $p^{\max} = 8$ dBm. 
For $(w_1=0.9, w_2 = 0.1)$ (the red line) in Fig.~\ref{fig:total_e_t_different_p}(b), it stresses the importance of energy, so it tries to find a smaller weighted sum of energy and time consumption, especially for the energy. Thus, after $p^{\max}=8$ dBm, its time cost is higher than the benchmark, but the energy consumption is much smaller than the benchmark.

Fig. \ref{fig:total_e_t_different_f} compares the total energy consumption and completion time at different maximum CPU frequencies. Obviously, with the increase of maximum CPU frequency, the energy consumption of the benchmark algorithm is rising, which can be seen in Fig. \ref{fig:total_e_t_different_f}(a). Intuitively, if the CPU frequency is set at the maximum, the time consumption will be reduced. Therefore, in Fig. \ref{fig:total_e_t_different_f}(b), the total time consumption is decreasing as CPU frequency increases. However, the benchmark algorithm performs  better than our algorithm only when $w_1=0.9$ and $w_2 = 0.1$ after the maximum CPU frequency$=0.7$ GHz, because our algorithm mainly focuses on optimizing the energy when ($w_1 = 0.9$, $w_2 = 0.1$). Hence, our energy consumption is the smallest in Fig. \ref{fig:total_e_t_different_f}(a).

Besides, three lines related to our algorithm in Fig. \ref{fig:total_e_t_different_f} all enter a stationary phase after a certain maximum CPU frequency, because the optimal CPU frequency is already found for each device in the corresponding range of CPU frequency. Thus, even if increasing the maximum CPU frequency, the algorithm will find the previous optimal CPU frequency.

{
\subsubsection{Different $\rho$ with fixed $(w_1, w_2)$}
We fix the weight parameters $(w_1, w_2)=(0.5, 0.5)$ of energy and time and choose $\rho=1,~10,~20,~30,~40,~50,~60$, respectively, to compare total energy and time consumption at different maximum computational frequencies and transmission power limits. The results can be seen in Fig. \ref{fig:total_energy_diff_rho}. \textbf{MinPixel} is the same as the benchmark algorithm in Fig. \ref{fig:total_e_t_different_p} and Fig. \ref{fig:total_e_t_different_f}. \textbf{RandPixel} randomly assigns the video frame resolution for each user.

We plot Fig. \ref{fig:total_energy_diff_rho} by using logarithmic scale. It can be observed that in Fig. \ref{fig:total_energy_diff_rho}, both lines of energy consumption and time cost of $\rho=1$ and $\rho=5$ are below the line of MinPixel.
There is a noticeable gap between our algorithm and MinPixel.
In Fig. \ref{fig:total_energy_diff_rho}, energy consumption does not change much versus the maximum transmission power.
Since we optimize the sum of energy and time, though energy consumption does not change much, the total time decreases versus the maximum transmission power as shown in Fig. \ref{fig:total_energy_diff_rho}(b).
Compared to MinPixel, our algorithm ($\rho=1$) reduces around 85\% of the total energy consumption and 42\% of the total time. 
From $\rho=10$, the energy and time consumption increase significantly. 
\textcolor{black}{We analyze that this is caused by the increase of $\rho$.}
Larger $\rho$ means we emphasize the optimization of model accuracy. As $\rho$ increases, the selection of a higher resolution results in an increase in the proportion of computational energy and time to total energy and time.  That is why the lines of $\rho=10,~20,~30,~40,~50$ are above MinPixel.
The overlapping lines of $\rho=20,~30,~40,~50$ indicate that for most devices, those lines make the selection of $\overline{\text{s}}_{M}$.
Nevertheless, our algorithm still improves about 67\% of the total energy consumption and 38\% of the total time, compared to RandPixel.

\textcolor{black}{
Fig. \ref{fig:acc_cifar_mnist} and Fig. \ref{fig:rho_acc} show the relationship between $\rho$ and the average model accuracy. We implement the experiments on a workstation with three NVIDIA RTX 2080 Ti GPUs.
The purpose of the experiments is to evaluate the performance of image classification and object detection in relation to the choices of the video frame resolution by setting different $\rho$. The number of clients in Fig. \ref{fig:acc_cifar_mnist} is $10$. In Fig. \ref{fig:acc_cifar_mnist}(a), we set three resolution choices, which are $\{12^2, 20^2, 28^2\}$ pixels, for MNIST. 
The model we use is\footnote{MNIST is a dataset with hand-written digit images including $60000$ gray training images and $10000$ test images. The CNN has two $5\times5$ convolutional layers, a fully connected (fc) layer with $320\times50$ units, another fc layer with $50\times10$ units followed by a ReLU activation function and a log softmax activation function. The first convolutional layer is with $10$ channels, and the second is with $20$ channels. Both are followed by $2\times2$ max pooling and ReLU activation.} CNN. 
The number of local epochs and global communication rounds is $2$ and $10$, respectively. Each client has $6000$ images by default.
Besides, in Fig. \ref{fig:acc_cifar_mnist}(b), we set the options of the resolution are $\{8^2, 16^2, 32^2\}$ pixels for CIFAR-10, and the model is also\footnote{CIFAR-10 is a dataset with $60000$ color images and $10$ classes. The CNN architecture consists of two $5\times5$ convolutional layers, each of which is followed by ReLU activation and a 2D max pooling layer. After the convolutional layers, the first fc layer has $400$ inputs units and $120$ output units, the second fc layer has $120$ input and $84$ output units, and the third fc layer has $84$ input and $10$ output layers followed by ReLU activation and a log softmax activation function to produce the final output.} CNN. We set the local epochs as $5$, the global communication rounds as $200$ and the number of images on each client as $5000$. 
 Also, we simulate the non-IID\footnote{Note that non-IID issue is also a worthwhile research problem within FL. However, the focus of this paper is not on this issue. Some FL papers have emphasized and conducted in-depth research on the non-IID problem \cite{9528995, zhao_ensemble_2023}.} condition for both datasets. ``non-IID (1-class)'' means that each client only has data with one single label. ``non-IID (2-class)'' implies that each client has data with two kinds of labels. The unbalanced condition is simulated by randomly allocating the number of samples to each client.
As $\rho$ increases, it could be observed that the model accuracy is increasing. The resolution of images does have an impact on the model performance. Moreover, the performance of 1-class non-IID and unbalanced is the worst in both Fig. \ref{fig:acc_cifar_mnist}(a) and Fig. \ref{fig:acc_cifar_mnist}(b). Compared with non-IID, the impact of the unbalanced dataset is smaller. The effect of unbalanced MNIST on the model is even smaller because MNIST is a simple task. Furthermore, the image resolution has a greater impact on the accuracy of CIFAR-10 compared with MNIST since CIFAR-10 is more complex.
}

Fig. \ref{fig:rho_acc} is about the relationship between $\rho$ and the average object detection model accuracy. \textcolor{black}{We implement YOLOv5m~\cite{glenn_jocher_2022_7002879} and YOLOv3-tiny~\cite{redmon2018yolov3}}, which belong to the object detection architectures You Only Look Once (YOLO), on COCO dataset~\cite{lin2014microsoft}. \textcolor{black}{The FL algorithm we use is FedAvg \cite{mcmahan2017communication}}.
\textcolor{black}{YOLOv5m and YOLOv3-tiny are modified to resize video frames incoming to different resolutions}, and object detection is then conducted based on the resized frames. 
\begin{figure*}[t!]
    \centering
    \begin{minipage} {.48\textwidth}
    \includegraphics[width=0.9\linewidth]{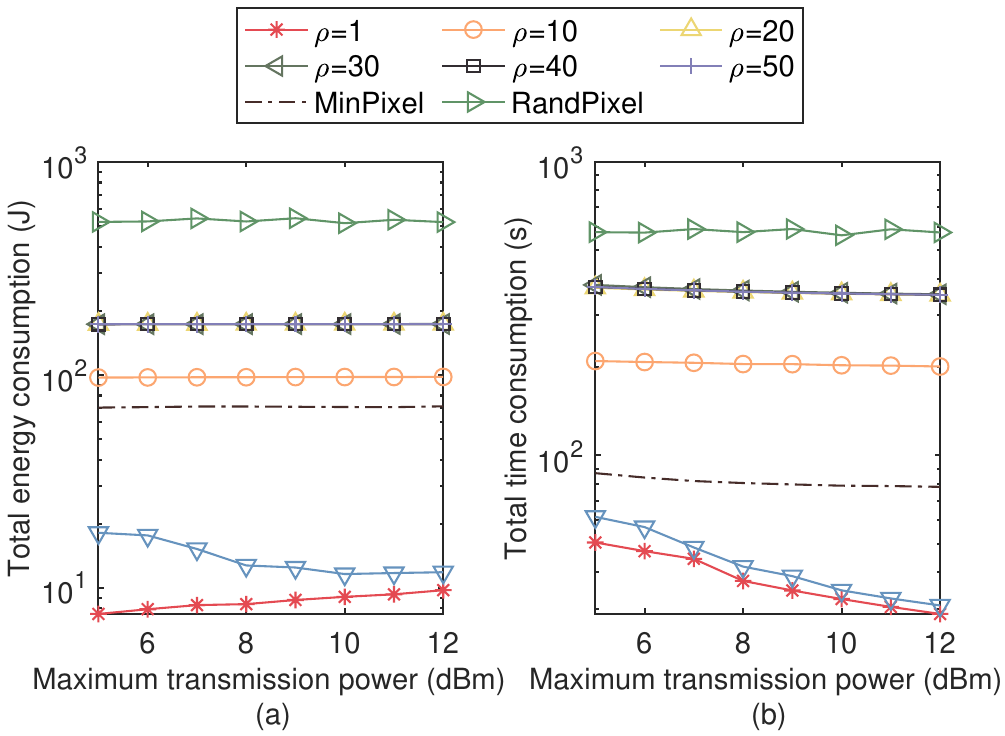}
    \caption{(a) Total energy consumption with different $\rho$ under different maximum transmit power limits; (b) Total time with different $\rho$ under different maximum transmit power limits. Remark: $(w_1, w_2)=(0.5, 0.5)$.}
    \label{fig:total_energy_diff_rho}
    \end{minipage}
    \begin{minipage} {.48\textwidth}
    \captionsetup{labelfont={color=black}}
        \includegraphics[width=0.98\linewidth]{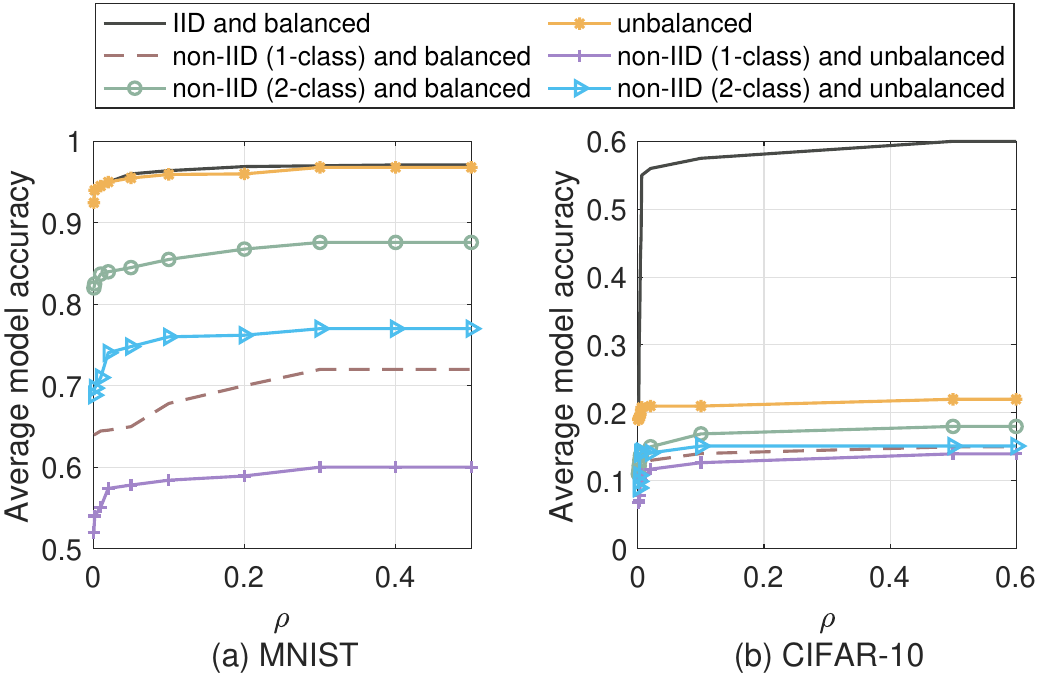}
    \caption{\color{black}The model accuracy of MNIST and \mbox{CIFAR-10} under different $\rho$. (a) The resolution options are $\{12^2, 20^2, 28^2\}$ pixels. (b) The resolution options are $\{8^2, 16^2, 32^2\}$ pixels. Remark: $(w_1, w_2)=(0.5, 0.5)$.}
    \label{fig:acc_cifar_mnist}
    \end{minipage}
\end{figure*}
Unlike the abovementioned setting, we simulate $20$ clients with one aggregation server. The local epoch $R_l$ is \textcolor{black}{$4$}, and the total aggregation round $R_g$ is $50$. \textcolor{black}{Each client has $5000$ samples by default. The options of the resolution are $\{160^2, 320^2, 480^2, 640^2\}$ pixels. For the unbalanced setting, we randomly allocate the number of samples to each client. We do not simulate the non-IID setting for this dataset because each image has several object categories. It is difficult to select images with one single label to simulate a pure non-IID environment. The batch size is set as 32. Here the model accuracy refers to the mean Average Precision, and it is a common metric for evaluating the model performance in the field of object detection.}
\textcolor{black}{Fig. \ref{fig:rho_acc} shows that, at first, all devices choose $\overline{\text{s}}_{1}=160$ as the video frame resolution, and thus the accuracy is the lowest. At around $\rho=15$, most devices will choose $s_n=320$ as their video frame resolution. 
As $\rho$ increases and when it reaches about $26$, devices begin to choose $s_n=480$ as their video frame resolution. With the continuous growth of $\rho$, at around $37$, devices gradually start choosing a resolution of $640$ pixels, with all devices selecting this resolution at approximately 40. This results in maximum accuracy, which is maintained thereafter. Note that YOLOv3-tiny is a smaller model, so it has a worse performance. Moreover, the unbalanced setting does have a negative impact on the model performance. However, compared to the impact of the resolution, its impact is relatively smaller.
}

\begin{figure*}[t!]
    \centering
    \begin{minipage} {.3\textwidth}
        \captionsetup{labelfont={color=black}}
    \centering
    \includegraphics[width=0.9\linewidth]{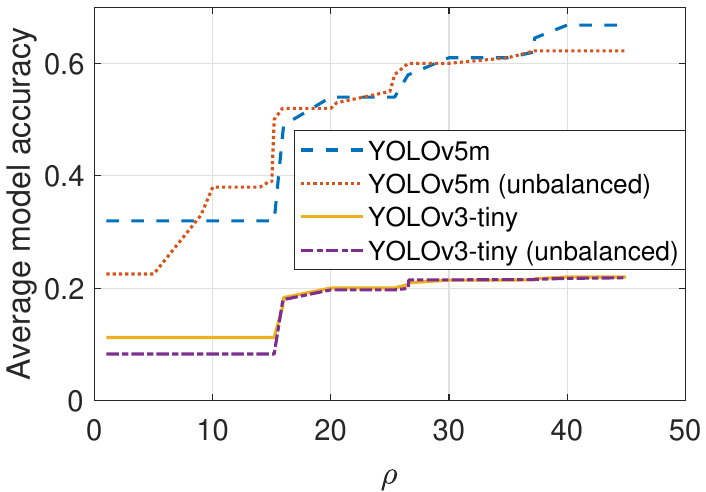}
    \caption{\color{black}Model accuracy under different $\rho$. The object detection is implemented on the COCO dataset using modified YOLO algorithm. Here $(w_1, w_2)=(0.5, 0.5)$.}
    \label{fig:rho_acc}
    \end{minipage}\hspace{4pt}
    \begin{minipage} {.27\textwidth}
    \centering
    \includegraphics[width=0.9\linewidth]{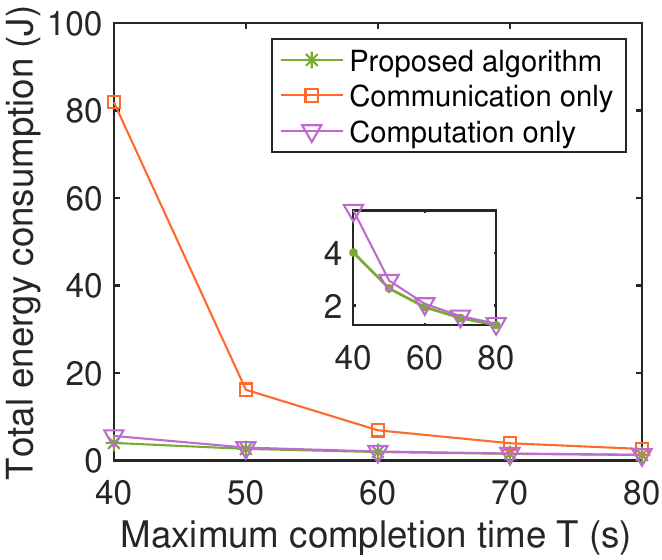}
    \caption{Total energy consumption versus maximum completion time. Here $(w_1, w_2)=(0.99, 0.01)$.}
    \label{fig:joint_commu_comp_comparison}
    \end{minipage} \hspace{2pt}
    \begin{minipage} {.35\textwidth}
    \centering
    \psfrag{T}{$\mathcal{T}$}
    \includegraphics[width=0.8\linewidth]{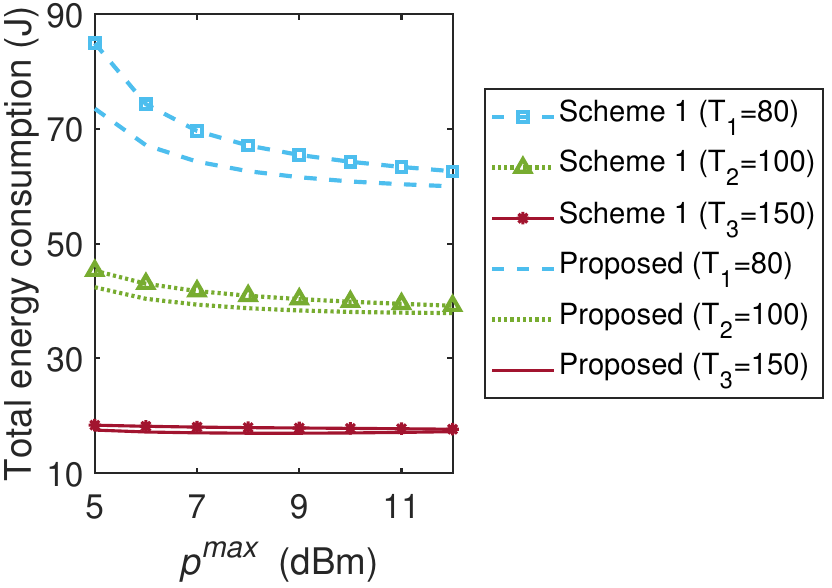}
    \caption{Total energy consumption versus maximum transmit power with fixed maximum completion time $T_1 = 80$ s, $T_2 = 100$ s and $T_3 = 150$ s. Here $(w_1, w_2)=(0.99, 0.01)$. Total number of users is $N = 50$.
    }
    \label{fig:comparison_e}
    \end{minipage}
\end{figure*}

}

\subsection{Joint Communication \& Computation Optimization vs. Communication/Computation Optimization Only}

To study the domination relationship between the transmission energy and computation energy, we compare our algorithm with the following schemes at the aspect of total energy consumption under different maximum completion times. The maximum transmission power is $p^{\max} = 10$ dBm. Because the maximum completion time $\mathcal{T}$ is fixed to compare these three schemes, we set $w_1=0.99$ and $w_2 = 0.01$ in our algorithm\footnote{Note that $w_2$ cannot be set to $0$ due to constraints (\ref{relation_f_s_lamda}) and (\ref{relation_w2_lamda}). Hence, we set $w_2=0.01$ to minimize the proportion of delay as much as possible.}.
\begin{itemize}
    \item \textbf{Communication optimization only}: This scheme only optimizes communication energy cost. Each device's CPU frequency is set as a fixed value, and the choice of video frame resolution is random.
    We only optimize the transmission power and bandwidth allocated to each device. To guarantee there is a feasible solution, the CPU frequency value is fixed for each device as $\frac{R_gR_l\zeta sn^2 c_nD_n}{T-R_g\max(d_n/r_n)}$, which is derived from constraint (\ref{constra:time}), and $r_n$ is calculated from the initial bandwidth and transmission power.
    \item \textbf{Computation optimization only}: This scheme only optimizes computation energy cost. Each device's transmission power and bandwidth are fixed. The transmission power and bandwidth of device $n$ are set as $p_n = p^{\max}$ and $B_n = \frac{B}{N}$.
\end{itemize}

Apparently, it can be observed from Fig. \ref{fig:joint_commu_comp_comparison} that our proposed algorithm performs better than the other two schemes. In addition, transmission energy takes the lead in total energy consumption. Only optimizing the computation energy cannot exceed the performance of only optimizing the transmission energy. Furthermore, we can discover the conflicting relationship between maximum completion time and total energy consumption. As maximum completion time increases, there will be better solutions to reduce total energy consumption. However, if we relax the constraint of the total delay $\mathcal{T}$ too much, the optimal solution will reach the minimum limit of transmission power and CPU frequency. That is why the gap between the lines of Computation-only and our proposed algorithm becomes small when $\mathcal{T}$ is large.

\subsection{Superiority of Our Optimization Algorithm over Existing Work} \label{secSuperiority}
Yang \emph{et~al.} \cite{yang2020energy} also proposed an optimization algorithm for the FL system via FDMA. We name their method Scheme 1 in our paper. Although Scheme 1 also has local accuracy as one of its optimization variables, its local accuracy refers to the accuracy of stochastic gradient descent algorithm and it is different from ours. Also, the optimization of the local accuracy is a separate section in Scheme 1. 
Thus, it will not affect the comparison between our method and Scheme 1. 
\textcolor{black}{Besides, Scheme 1 does not include the optimization of video frame resolution $s_n$, so we only compare the performance of our proposed algorithm without the optimization of video frame resolution (i.e., the proposed algorithm in our conference version \cite{ICDCS}) and Scheme 1 in terms of energy and latency optimization. Hence, we only use the method (Algorithm 3 in \cite{yang2020energy}) they use to optimize the transmission power, the bandwidth and the CPU frequency of each user.}

The maximum completion time is not included in the objective function of the optimization problem in Scheme~1. Instead, it appears as a constraint. Therefore, to be fair, we set a fixed total completion time $\mathcal{T}$ to compare our optimization algorithm with Scheme 1. With a fixed maximum completion time $\mathcal{T}$, our minimization problem (\ref{equa:min2}) is under the condition that $w_1 = 0.99$ and $w_2 = 0.01$. Besides, the initial transmission power and bandwidth of device $n$ are set as $p_n = p^{\max}$, $B_n = \frac{B}{2N}$. The comparison result is shown in Fig. \ref{fig:comparison_e}.

From Fig. \ref{fig:comparison_e}, we can find that three green lines are below the corresponding black lines, which reveals that our algorithm's performance is better. Additionally, as the maximum completion time $\mathcal{T}$ decreases, the total energy consumption gap between our algorithm and Scheme 1 gets larger, thereby indicating that our proposed algorithm leads to a better solution when the maximum completion time $\mathcal{T}$ is small. Therefore, our proposed algorithm will be more appropriate for the training scenario with a restricted training time.

At $T_3 = 150$ s, the solid line representing our algorithm first decreases and then increases. This is because our initial transmission power is set as $p_n = p^{\max}$. After finding the optimal solution at a specific $p^{\max}$, the subsequent total energy consumption will slightly increase since the different starting point from the previous step as $p^{\max}$ continues increasing.

\section{Conclusion} \label{sec:conclusion}
Our paper considers an optimization problem for the FL system with MAR for the Metaverse. The optimization variables include the bandwidth allocation, transmission power, CPU frequency setting and MAR video frame resolution of each participating device. By introducing three weight parameters, we optimize a weighted combination of the total energy consumption, completion time and model accuracy. By adjusting the weight parameters, our resource allocation scheme can be amenable to different requirements of the FL-MAR system for the Metaverse. We present an algorithm to solve the optimization problem via a rigorous analysis. The convergence and computational complexity of the proposed algorithm are also provided. In the experiments, we observe that our resource allocation algorithm outperforms the existing schemes.

\renewcommand{\refname}{~\\[-30pt]References }



\renewcommand{\theequation}{A.\arabic{equation}}

  \setcounter{equation}{0}

\begin{appendices}

\section{Notations}\label{notations}

\begin{table}[H] 
\centering
\caption{Notations}
\label{tab:notation}
\begin{tabular}{|c|c|}
\hline
\textbf{Description}                                        & \textbf{Notation}            \\ \hline
The number of total devices                                 & $N$                          \\ \hline
Total energy consumption                                    & $\mathcal{E}$                \\ \hline
\begin{tabular}[c]{@{}l@{}}The completion time of the whole \\ federated learning process\end{tabular} & $\mathcal{T}$                \\ \hline
The accuracy metric                                         & $\mathcal{A}$                \\ \hline
Maximum uplink bandwidth                                    & $B$                          \\ \hline
Video frame resolution of device $n$                        & $s_n$                        \\ \hline
Allocated bandwidth of device $n$                           & $B_n$                        \\ \hline
The number of samples on device $n$                         & $D_n$                        \\ \hline
Total number of samples on all devices                      & $D$                          \\ \hline
Transmission data size of user device $n$                   & $d_n$                        \\ \hline
\begin{tabular}[c]{@{}l@{}}Maximum completion time between\\ two global aggregations\end{tabular}     & $T$                          \\ \hline
Data transmission time of device $n$                        & $T^{trans}_n$                   \\ \hline
Local computation time of user device $n$                   & $T^{cmp}_n$                  \\ \hline
Data transmission rate of device $n$                        & $r_n$                        \\ \hline
Minimum transmission rate of device $n$                     & $r_n^{\min}$                 \\ \hline
CPU frequency of device $n$                                 & $f_n$                        \\ \hline
\begin{tabular}[c]{@{}l@{}}Minimum and maximum CPU\\ frequency of device $n$    \end{tabular}         & $f_n^{\min}, f_n^{\max}$     \\ \hline
\begin{tabular}[c]{@{}l@{}}The number of local iterations\\in one global round \end{tabular}         & $R_l$                        \\ \hline
The number of global rounds                                 & $R_g$                        \\ \hline
The  effective switched capacitance                         & $\kappa$                     \\ \hline
CPU cycles per standard sample of device $n$                & $c_n$                        \\ \hline
\begin{tabular}[c]{@{}l@{}}Local computation energy consumption\\ of user device $n$  \end{tabular}   & $E_n^{cmp}$                  \\ \hline
Transmission energy consumption of device $n$               & $E_n^{trans}$                \\ \hline
Weight parameters                                           & $w_1$, $w_2$, $\rho$         \\ \hline
Noise power spectral density                                & $N_0$                        \\ \hline
Transmission power of device $n$                            & $p_n$                        \\ \hline
\begin{tabular}[c]{@{}l@{}}Minimum and maximum transmission\\ power of device $n$   \end{tabular}     & $p_n^{\min}, p_n^{\max}$     \\ \hline
Channel gain from device $n$ to the base station            & $g_n$                        \\ \hline
Lagrange multipliers                                        & $\bm{\lambda},\bm{\tau},\mu$ \\ \hline
Auxiliary variables                                         & $\bm{\nu},\bm{\beta}$        \\ \hline 
The number of resolution choices     & $M$ \\ \hline
\end{tabular}
\end{table}

\section{Solution to Subproblem 1} \label{sol_to_sp1}
We treat the optimization problem by the Karush--Kuhn--Tucker (KKT) approach
and introduce the Lagrange function
\begin{align} \label{lag_sp1}
    &L_1(f_n, \hat{s_n}, T, \bm{\lambda}) \!=\! w_1R_g \sum_{n=1}^N \kappa R_l\zeta \hat{s_n}^2 c_n D_n f_n^2 \!+\! w_2R_g  T \notag \\
    &\!-\! \rho\sum_{n=1}^N  \hat{A}_n(\hat{s_n}) 
    \!+\! \sum_{n=1}^N \lambda_n [ (\frac{R_l\zeta \hat{s_n}^2c_nD_n}{f_n} \!+\! T_n^{trans}) \!-\! T],
\end{align}
where $\bm{\lambda}:=[\lambda_1, \lambda_2, \ldots, \lambda_N]$ is the Lagrange multiplier associated with the inequality constraint (\ref{constra:time}). 

After applying KKT conditions to (\ref{lag_sp1}), we get 
\begin{align}
    & \frac{\partial L_1}{\partial f_n} \!=\! 2w_1 R_g R_l\kappa \zeta \hat{s_n}^2c_nD_nf_n \!-\! \lambda_n\frac{R_l\zeta \hat{s_n}^2c_nD_n}{f_n^2} \!=\! 0 , \label{KKT:f_n} \\
    &{ 
    \frac{\partial L_1}{\partial \hat{s_n}} = 2w_1R_g R_l\kappa \zeta \hat{s_n}c_nD_nf_n^2 + \lambda_n\frac{2R_l\zeta \hat{s_n}c_nD_n}{f_n}} \notag \\
    &~~~~~-\rho A^{\prime}_n(\hat{s}_n)=0,\label{KKT:s_n} 
    \end{align}
\begin{align}
    &\frac{\partial L_1}{\partial T} = w_2R_g-\sum_n^N\lambda_n = 0,\\
    &\lambda_n [ (\frac{R_l\zeta\hat{s_n}^2c_nD_n}{f_n}+T_n^{trans})-T] = 0,
\end{align}
from which we derive 
\begin{align}
    f_n^* &= \sqrt[3]{\frac{\lambda_n}{2w_1R_g\kappa}},~\hat{s_n}^*= \frac{\rho A^{\prime}_n(\hat{s_n}^*)}{2R_l\zeta c_nD_n \cdot [w_1R_g\kappa (f_n^*)^2 + \frac{\lambda_n}{f_n^*}]}, \label{relation_f_s_lamda}\\[-5pt]
    w_2 &= \frac{\sum_n^N \lambda_n}{R_g}.\label{relation_w2_lamda}
\end{align}

\textcolor{black}{Here, we discuss the special case that $A_n(\hat{s_n})$ is linear and obtain its dual problem with only one variable to reduce the time complexity. Suppose $A_n(\hat{s_n})=\hat{k_n}(\hat{s_n}-\overline{\text{s}}_1)+\text{A}_{\overline{\text{s}}_1}$, where $\hat{k_n}=\frac{\text{A}_{\overline{\text{s}}_1}-\text{A}_{\overline{\text{s}}_M}}{\overline{\text{s}}_M-\overline{\text{s}}_1}$, and $\text{A}_{\overline{\text{s}}_1}$ and $\text{A}_{\overline{\text{s}}_M}$ represent the accuracy of corresponding resolutions $\overline{\text{s}}_1$ and $\overline{\text{s}}_M$, respectively.} Through using Eq. (\ref{relation_f_s_lamda}), we can use $\lambda_n$ to represent $f_n$ and $\hat{s_n}$. Then, the dual problem is 
\begin{subequations} \label{equa:subproblem1_dual}
\begin{align} 
\max_{\lambda_n}~&\sum_{n=1}^N -\frac{\rho^2\hat{k}_n^2}{4h(2^{-\frac{2}{3}} + 2^{\frac{1}{3}})}\lambda_n^{-\frac{2}{3}} + T_n^{trans}\lambda_n  +\rho\hat{k}_n \overline{\text{s}}_{1}-\rho \text{A}_{\overline{\text{s}}_{1}} \tag{\ref{equa:subproblem1_dual}} \\[-5pt]
\text{subject to}, \notag \\[-5pt]
&\sum_{n=1}^N \lambda_n =  w_2R_g,~\lambda_n \geq 0, \label{constra:lambda_w2}
\end{align}
\end{subequations}
where $h = R_l\zeta c_nD_n(w_1\kappa R_g)^{\frac{1}{3}}$. Obviously, this dual problem is a simple convex optimization problem. In this paper, we use CVX \cite{grant2014cvx} 
to solve it and get the optimal $\bm{\lambda^*} = [\lambda_1^*,..., \lambda_n^*]$. Then, we are able to calculate $\bm{f}$ and $\bm{\hat{s}}$ through Eq. (\ref{relation_f_s_lamda}).

\section{Proof of Lemma \ref{lemma:requisite_fra_prog}} \label{appen:proof_lemma_requis}
It is obvious that $K(p_n, B_n) = p_nd_n$ is an affine function with $B_n=0$. Thus, $K(p_n, B_n)$ is convex.

To prove $G(p_n, B_n)$ is concave, we first calculate its Hessian.
\begin{align}
    Hessian(G) =  
    \begin{bmatrix}
      -\frac{g_n^2}{B_nN_0^2(\frac{g_np_n}{B_nN_0}+1)^2} & \frac{g_n^2p_n}{B_n^2N_0^2(\frac{g_np_n}{B_nN_0} + 1)^2} \\
      \frac{g_n^2p_n}{B_n^2N_0^2(\frac{g_np_n}{B_nN_0} + 1)^2} & -\frac{g_n^2p_n^2}{B_n^3N_0^2(\frac{g_np_n}{B_nN_0} + 1)^2}
    \end{bmatrix}.
\end{align}

Given a vector $\bm{x}=[x_{1}, x_{2}]^T$ in $\mathbb{R}^{2}$, we have
\begin{align}
   \bm{x}^THessian(G)\bm{x} = -\frac{(x_{1}g_nB_n-x_{2}g_np_n)^2}{B_n^3N_0^2(\frac{g_np_n}{B_nN_0}+1)^2} \le 0.
\end{align}
Because $\bm{x}^T*Hessian(G)*\bm{x} \le 0$ for all $\bm{x}$ in $\mathbb{R}^{2}$, $Hessian(G)$ is a negative semidefinite matrix. Therefore, $G(p_n, B_n)$ is a concave function. \textbf{Lemma \ref{lemma:requisite_fra_prog}} is proved.

\section{Proof of Theorem \ref{theor:express_B_p}} \label{appen:proof_express_bp}
Before applying KKT conditions, we write down the partial Lagrangian function of problem \textbf{\textit{SP2\_v2}}:  
\begin{align} 
    &L_2(p_n, B_n, \tau_n, \mu)= \sum_{n=1}^N \nu_n(p_nd_n-\beta_nB_n\log_2(1+\frac{p_ng_n}{N_0B_n})) \notag\\
    &- \sum_{n=1}^N \tau_n(B_n\log_2(1 +\frac{p_ng_n}{N_0B_n})-r_n^{\min}) + \mu(\sum_{n=1}^NB_n - B),\label{equal:sp2_v2_lagran}
\end{align}
where $\tau_n|_{n=1,\ldots,N}$ and $\mu$ are non-negative Lagrange multipliers.

After applying KKT conditions to problem \textbf{\textit{SP2\_v2}}, we get
\begin{align}
    \frac{\partial L_2}{\partial p_n} & = \nu_n(d_n-\frac{\beta_ng_n}{N_0(1+\vartheta_n)\ln2})-\frac{\tau_ng_n}{N_0(1+\vartheta_n)\ln2} =0, \label{lag:partial_p}\\[-5pt]
    \frac{\partial L_2}{\partial B_n} &= -(\nu_n\beta_n+\tau_n)\log_2(1+\vartheta_n)
    +\frac{(\nu_n\beta_n+\tau_n)p_ng_n}{(1+\vartheta_n)\ln2 N_0B_n}  \notag\\
    &+\mu = 0,  \label{lag:partial_B}\\
    -\tau_n & (B_n\log_2(1+\vartheta_n)-r_n^{\min}) = 0, \label{lag:data_rate}\\[-5pt]
    \mu (\sum_{n=1}^N & B_n-B) = 0, \label{lag:mu}
\end{align}
where $\vartheta_n = \frac{p_ng_n}{N_0B_n}$ for $n=1,\cdots,N$.

We derive the relationship between $\bm{p}$ and $\bm{B}$ from (\ref{lag:partial_p}), so 
\begin{align} \label{equa:p_B_relation}
    p_n = (\frac{(\nu_n\beta_n+\tau_n)g_n}{N_0d_n\nu_n\ln2}-1)\frac{N_0B_n}{g_n}
\end{align}
From (\ref{lag:data_rate}) and (\ref{equa:p_B_relation}), we could analyze
\begin{align}
    B_n = \frac{r_n^{\min}}{\log_2(\frac{(\nu_n\beta_n+\tau_n)g_n}{N_0d_n\nu_n\ln2})}, \text{~if~} \tau_n \neq 0. \label{smallest_Bn}
\end{align}
Substituting (\ref{equa:p_B_relation}) in (\ref{equal:sp2_v2_lagran}), we replace $p_n$ by $B_n$ in the Lagrangian function:
\begin{align}
&L_3(B_n,  \tau_n, \mu) = \big(\frac{\nu_n\beta_n+\tau_n}{\ln2}-\frac{\nu_nd_nN_0}{g_n}-(\nu_n\beta_n+\tau_n)\notag\\
&\times   \log_2(\frac{(\nu_n\beta_n+\tau_n)g_n}{N_0d_n\nu_n\ln2})+\mu\big)B_n+\sum_{n=1}^N r_n^{\min}* \tau_n-\mu B.
\end{align}
The corresponding dual problem writes as
\begin{align} \label{equa:subp2_dual}
\max_{\tau_n, \mu} ~& g(\tau_n, \mu) = \sum_{n=1}^N r_n^{\min}*\tau_n-\mu B\\[-5pt]
\text{subject to}~& \frac{a_n}{\ln2}-j_n-a_n\log_2(\frac{a_n}{j_n\ln2})+\mu = 0, \label{subp2_constra1}\\[-5pt]
&\mu \ge 0, ~\tau_n \ge 0,
\end{align}
where $a_n = \nu_n\beta_n+\tau_n$ and $j_n = \frac{\nu_nd_nN_0}{g_n}$.
Besides, constraint (\ref{subp2_constra1}) derives the relationship between Lagrange multipliers $\tau_n$ and $\mu$: 
\begin{align}\tau_n = \frac{(\mu-j_n)\ln2}{W(\frac{\mu-j_n}{e*j_n})}-\nu_n\beta_n, \label{tau_mu_relation}
\end{align}
where $W(\cdot)$ is Lambert $W$ function and $e$ is the base of the natural logarithms.

Given (\ref{tau_mu_relation}), the dual function is simplified as 
\begin{align}
g(\mu) = \sum_{n=1}^N r_n^{\min}*(\frac{(\mu-j_n)\ln2}{W(\frac{\mu-j_n}{e*j_n})}-\nu_n\beta_n)-\mu B.
\end{align}
Take the first derivative and we get $g'(\mu) = \sum_{n=1}^N\frac{r_n^{\min}\ln2}{W(\frac{\mu-j_n}{e*j_n})}(1-\frac{1}{W(\frac{\mu-j_n}{e*j_n})+1})-B$.
Additionally, $g''(\mu)=-\sum_{n=1}^N \frac{r_n^{\min}\ln2W(\frac{\mu-c_n}{e*c_n})}{(\mu-j_n)(1+W(\frac{\mu-j_n}{e*j_n}))^3} \le 0$ because $\frac{\mu-j_n}{e*j_n}\ge -\frac{1}{e}$.

Therefore, $g'(\mu)$ is a monotone decreasing function, and $g(\mu)$ is concave, so $g(\mu)$ reaches maximum when $g'(\mu) = 0$. The bisection method can be used to find $\mu$ satisfying $g'(\mu) = 0$. Naturally, $\tau_n = \max((\ref{tau_mu_relation}), 0)$.
Note $\tau_n \neq 0$ implies $B_n = \frac{r_n^{\min}}{\log_2(\frac{(\nu_n\beta_n+\tau_n)g_n}{N_0d_n\nu_n\ln2})}$. We denote the sum of bandwidth of these devices by $B_{\tau_n\neq0}$ and the number of these devices by $N_{\tau_n\neq0}$

Then, replacing $p_n$ by $B_n$ according to (\ref{equa:p_B_relation}) in problem \textbf{\textit{SP2\_v2}}, removing those devices whose bandwidth is calculated in the previous step and considering the remaining devices, the new problem becomes
{
\begin{align} \label{SP2_v3}
  \min_{B_n}~~~~&\hspace{-20pt}\sum_{n=1}^{N-N_{\tau_n \neq 0}}  (\frac{\nu_n\beta_n}{\ln2}-\frac{N_0d_n\nu_n}{g_n}-\nu_n\beta_n \times  \log_2(\frac{\beta_ng_n}{N_0d_n\ln2}))B_n \\[-5pt]
 \text{subject to}, \notag \\[-5pt]
& \hspace{-25pt}p_n^{\min} \le (\frac{\beta_ng_n}{N_0d_n\ln2}-1)\frac{N_0B_n}{g_n} \le p_n^{\max}, ~\forall n   \in \mathcal{N}\backslash\mathcal{N}_{\tau_n\neq0}, \\[-5pt]
& \sum_{n=1}^{N-N_{\tau_n\neq0}} B_n \le B-B_{\tau_n\neq0},
\end{align}}
where $\mu$ and $\tau_n$ have already been solve in previous steps.

Until now, the problem \textbf{\textit{SP2\_v2}} has become a more simple convex optimization problem with just one variable. Thus, we can use the convex problem solver CVX \cite{grant2014cvx} to solve problem (\ref{SP2_v3}).
\end{appendices}

\end{document}